\documentclass[journal]{IEEEtran} 
\usepackage{pifont} 
\usepackage{amsmath,amsfonts,amssymb,amsthm,mathtools}
\usepackage{algorithmic}
\usepackage{algorithm}
\usepackage{array}
\usepackage[caption=false,font=footnotesize]{subfig}
\usepackage{textcomp}
\usepackage{stfloats}
\usepackage{url}
\usepackage{verbatim}
\usepackage{graphicx}
\usepackage{cite}
\usepackage{multirow, booktabs} 
\newcommand{\cmark}{\textcolor{green}{\ding{51}}}%
\newcommand{\xmark}{\textcolor{red}{\ding{55}}}%

\usepackage[most]{tcolorbox} 
\newtcbox{\mybox}[1][]{nobeforeafter,tcbox raise base,colframe=green!50!black,colback=green!10!white,top=0pt,bottom=0pt,left=0pt,right=0pt,before upper=\strut,#1}

\newcommand{\norm}[1]{\left\lVert#1\right\rVert}
\newcommand{\normsq}[1]{\left\lVert#1\right\rVert^2}
\newcommand{\expec}[1]{\mathbb{E}\left[#1\right]}
\newcommand{\expecq}[1]{\mathbb{E}_{Q}\left[#1\right]}

\newcommand{\order}[1]{\mathcal{O}\left(#1\right)}

\newcommand{\calS}{\mathcal{S}}

\DeclareMathOperator{\topk}{top-k}

\DeclareMathOperator{\qsgd}{QSGD}

\usepackage{hyperref} 
\usepackage[capitalize]{cleveref} 

\usepackage{tikz}
\usetikzlibrary{calc,shapes,arrows,positioning}
\usetikzlibrary{shapes.geometric}
\usetikzlibrary{arrows.meta}

\theoremstyle{plain}
\newtheorem{theorem}{Theorem}[section]

\newtheorem{lemma}[theorem]{Lemma}
\newtheorem{corollary}[theorem]{Corollary}
\theoremstyle{definition}
\newtheorem*{definition*}{Definition}

\newtheorem{assumption}[theorem]{Assumption}
\crefname{assumption}{Assumption}{Assumptions}

\theoremstyle{remark}
\newtheorem{remark}[theorem]{Remark}

\begin{document}
\bstctlcite{IEEEexample:BSTcontrol} 

\title{Quantized and Asynchronous Federated Learning}

\author{Tomas Ortega, \IEEEmembership{Graduate Student Member, IEEE}, and Hamid Jafarkhani, \IEEEmembership{Fellow, IEEE}
    \thanks{
        Authors are with the Center for Pervasive Communications \& Computing and  EECS Department, University of California, Irvine, Irvine, CA 92697 USA (e-mail: \{tomaso, hamidj\}@uci.edu). This work was supported in part by the NSF Award ECCS-2207457. An early version of this work was presented at the 2023 ICML workshop of Federated Learning and Analytics in Practice~\cite{QAFeLworkshop}.} }


\maketitle

\begin{abstract}
    Recent advances in federated learning have shown that asynchronous variants can be faster and more scalable than their synchronous counterparts.
    However, their design does not include quantization, which is necessary in practice to deal with the communication bottleneck.
    To bridge this gap, we develop a novel algorithm, Quantized Asynchronous Federated Learning (QAFeL), which introduces a hidden-state quantization scheme to avoid the error propagation caused by direct quantization.
    QAFeL also includes a buffer to aggregate client updates, ensuring scalability and compatibility with techniques such as secure aggregation.
    Furthermore, we prove that QAFeL achieves an $\order{1/\sqrt{T}}$ ergodic convergence rate for stochastic gradient descent on non-convex objectives, which is the optimal order of complexity, without requiring bounded gradients or uniform client arrivals.
    We also prove that the cross-term error between staleness and quantization only affects the higher-order error terms.
    We validate our theoretical findings on standard benchmarks.
\end{abstract}


\section{Introduction}

\IEEEPARstart{I}{n} a traditional machine learning (ML) pipeline, data is collected from clients at a central server.
Then, a model is trained on the collected data and is deployed for use.
This has two major drawbacks: (i) it requires a large amount of storage at a central server, and more importantly, (ii) it raises privacy concerns when collecting sensitive data.
Decentralized learning techniques can deal with these concerns \cite{D-ADMM,shen2021distributed,chocosgd}.
One of the mechanisms to inherently address these drawbacks is federated learning (FL) where clients train local models and send them to the server for aggregation~\cite{communication_efficient}.
In FL, client data is used exclusively to train local models, ensuring it never leaves the client side.
Instead, the clients send their local models' updates to the server and the server aggregates the updates to create a global model.
Including new updates improves the global model's accuracy.
FL is gaining momentum in healthcare, finance, and natural language processing, to name a few areas~\cite{advances_open_problems,tfl-dt}.

FL characteristics are different from those of the traditional distributed optimization.
First, the data originates from the clients and cannot be shared with the server.
Second, clients are heterogeneous, i.e., they have access to different amount of data and operate with different speeds and communication bandwidths~\cite{joint_TCOM}.
Since the size of ML models is large, and becoming larger with time (particularly with language models)~\cite{ML_model_sizes}, communicating model updates from edge devices with bandwidth constraints is a costly operation.

\subsection{Related work}
Multiple FL algorithms have been proposed in the literature such as FedSGD~\cite{communication_efficient}, FedAvg~\cite{fedavg_conv_Li}, and FedProx~\cite{fedprox}, to name a few.
In FedSGD, first, each client completes a stochastic gradient descent (SGD) step with its local data and sends the update to the server.
Then, the updated local models are averaged at the server to create a global model.
Finally, the global averaged model is sent back to clients and the process is repeated.
To reduce communication costs, FedAvg allows multiple SGD steps at each client before exchanging the model, proving effective in real-world scenarios.
FedProx builds upon FedAvg, introducing a proximal term to account for disparities in local data distributions and device characteristics.

\subsubsection{Synchronous vs. Asynchronous Federated Learning}
\begin{figure}[htbp]
    \centering
    \includegraphics[width=\columnwidth]{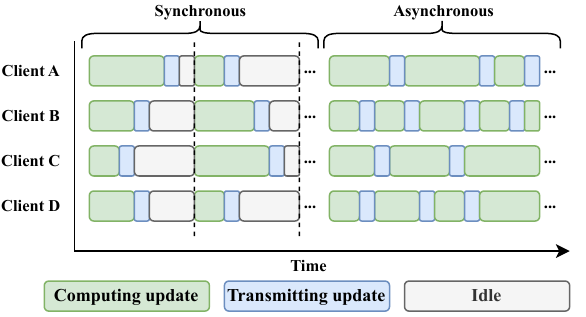}
    \caption{Flowchart comparing synchronous and asynchronous FL.}
    \label{fig:sync_vs_async}
\end{figure}

A common assumption is that these methods operate in rounds, leading to synchronous federated learning (FL), where the server waits for a predetermined time to receive updates from all clients.
Should a client miss the time window, its update is considered stale and is discarded.
While FedProx allows clients to perform different number of local steps, it still requires them to communicate before the round ends.
Since FL is designed for massive-scale networks, it is natural for clients to have different update times~\cite{async_hetero}.
As a result, interest has grown in asynchronous FL, which allows the server to update the global model without waiting for all clients.
There is no idle time in asynchronous FL and the clients restart calculating a new update after each transmission.
A comparison between synchronous and asynchronous FL is shown in \cref{fig:sync_vs_async}.

Despite its challenges, like the need to handle stragglers and stale gradients, asynchronous FL is attractive because it eliminates the burden of fitting clients into time slots.
This enables slow clients to participate in the training process and enables the use of larger training cohorts~\cite{async_edge,papaya}.
A naive approach to asynchronous FL is to have the server update the global model every time it receives a client update.
However, this method is not scalable, as the communication cost of broadcasting the global model grows too much with the number of clients.
To address this issue, client updates are buffered at the server before performing a global model update.
FedBuff~\cite{FedBuff,toghani2022unbounded}, an asynchronous and scalable version of FedAvg, is an example of such an algorithm suitable for heterogeneous clients.
It allows multiple local steps at each client and is compatible with privacy-preserving mechanisms.
It is also fairer and more efficient compared to synchronous FL methods~\cite{papaya}.

Other asynchronous FL works include algorithms that consider time-weighted schemes for stale model aggregation \cite{robust-async,backdoor-FL}. 
In another approach, \cite{zhou2022fedaca} proposes a client uploading policy, which avoids sending updates where the previous and current models are sufficiently similar.
However, these solutions do not consider quantization, which is indispensable to reduce the communication bottleneck in many practical FL systems.

\subsubsection{Quantization in Federated Learning}
\begin{figure}[htbp]
    \centering
    \includegraphics{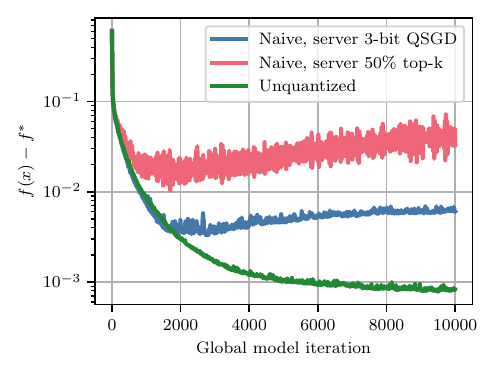}
    \caption{Numerical example of the effect of direct quantization (Naive) vs. the unquantized counterpart. No client quantization is performed. We show unbiased ($\qsgd$ \cite{qsgd}) and biased ($\topk$ \cite{error_feedback}) server quantizer examples.
        We consider a logistic regression problem with $\ell_2$ regularization on the \emph{mushrooms} dataset from LIBSVM \cite{LIBSVM}.
        The simulation parameters are: 100 clients, delays following a half-normal distribution, server buffer size of 10, client learning rate of 2, server learning rate of 0.1, and $\ell_2$ regularization strength of $1/8124$, where $8124$ is the number of samples in the dataset.
        The $y$-axis illustrates the difference $f(x) - f^\star$, where $f(x)$ is the global model cost at a given iteration, and $f^*$ is the optimal cost.
    }
    \label{fig:FedBuffvsNaive}
\end{figure}

Apart from allowing multiple local steps, quantization can further reduce the communication overhead.
There is a vast literature on synchronous FL with quantized communications \cite{UVeQFed, optimal_compression, error_feedback, unbiased_horvath}.
Quantization can reduce the number of transmitted bits in both directions: (i) the server can quantize the global model prior to sending it to clients, and (ii) clients can quantize the model updates before transmitting them to the server.
Moreover, quantizing client updates enhances the privacy guarantees~\cite{joint_privacy_quantization}.
On the other hand, directly quantizing the models results in error propagation over time.
For example, since clients only access the quantized global model, there will be a drift between the global models at the server and clients.
This phenomenon is illustrated in \cref{fig:FedBuffvsNaive}.
We observe that a naive use of quantization results in suboptimal loss.
This is particularly egregious in the biased quantization case, where FedBuff diverges when a server sends the top 50\% of coordinates in absolute value, setting the rest to zero.
This is a popular quantizer with a low compression ratio of $1/2$ and the fact that it diverges is devastating.

The use of quantization in asynchronous FL remains unexplored.
It is of particular interest to investigate the error produced by the compounded effects of model staleness due to asynchrony and quantization error, since the former is absent in the synchronous FL setting.

\subsubsection{Managing Error Propagation}
To manage the error propagation while quantizing the model, the server and clients should operate on the same model.
Therefore, we define a common model state and keep it at all nodes.
The difference between the updated server model and the common model state is quantized and communicated after every server update.
Similar ideas exist in other signal processing fields, for example in handling the drifts caused by motion compensation in  video coding \cite{error_feedback_video, video_codec_design,md-video-coding}.
In quantization theory, the general framework to manage error propagation is discussed under predictive coding \cite{quantization}.
There has been some efforts in managing error propagation in synchronous FL \cite{sparsifiedSGD,sparsifiedGradientMethods,EF21,chocosgd,ortegaGossip,MCM}, but, to the best of our knowledge, this is the first effort in handling the error propagation in asynchronous FL.

\subsection{Contributions}
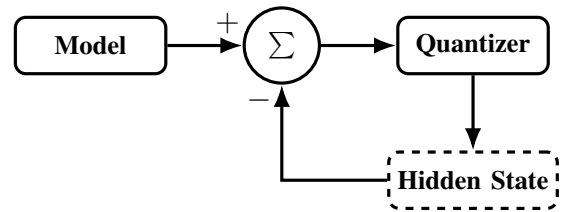
\begin{figure}[htbp]
    \centering
    \tikzstyle{edge} = [very thick, -{Latex}]
\tikzstyle{vedge} = [very thick, magenta, dashdotted, -{Latex}]
\tikzstyle{my-circle} = [draw, circle, fill=white, very thick, minimum width=1cm, minimum height=1cm]
\tikzstyle{rounded-rectangle} = [draw, rectangle, rounded corners, fill=white, very thick, minimum width=2cm, minimum height=.75cm]
\tikzstyle{dashed-rounded-rectangle} = [draw, rectangle, rounded corners, fill=white, very thick, dashed, minimum width=2cm, minimum height=.75cm]

\begin{tikzpicture}

    \node[my-circle] (sum) {\textbf{$\sum$}};
    \node[xshift=-0.3cm, yshift=-0.2cm] (minus) at (sum.south) {\large{$-$}};
    \node[xshift=-0.2cm, yshift=0.3cm] (minus) at (sum.west) {\large{$+$}};
    \node[rounded-rectangle] (state) [left = of sum] {\textbf{Model}};
    \node[rounded-rectangle] (quantizer) [right = of sum] {\textbf{Quantizer}};
    \node[dashed-rounded-rectangle] (hidden) [below = of quantizer] {\textbf{Hidden State}};

    \draw[edge] (state) -- (sum);
    \draw[edge] (sum) -- (quantizer);
    \draw[edge] (quantizer) -- (hidden);
    \draw[edge] (hidden) -| (sum);

\end{tikzpicture}
    \caption{Block diagram for updating the hidden-state.} 
    \label{fig:hidden-state-block-diagram}
\end{figure}

Motivated by the need to reduce communication overhead and the appeal of asynchronous FL, we propose a new algorithm, called Quantized Asynchronous Federated Learning (QAFeL), which includes a bidirectional quantization scheme for asynchronous FL with buffered aggregation.
To address the error propagation, we introduce a common hidden-state by aggregating all communicated messages as shown in \cref{fig:hidden-state-block-diagram}.
The server quantizes and broadcasts the difference between the hidden-state and its updated model.
Similarly, clients quantize the difference between their updated model and the corresponding hidden-state version.
Using the mechanism in \cref{fig:hidden-state-block-diagram}, QAFeL avoids error propagation and is scalable as it only needs to track one hidden-state.
Furthermore, it is a privacy-aware system since it does not track client states.
We also investigate the QAFeL's compound error produced by staleness and quantization to gain insights not present in separate analysis of the staleness and quantization effects.
\Cref{tab:comp} compares the characteristics of our work with respect to others.
\begin{table*}[htbp]
\centering
\caption{Related work comparison. The number of algorithm rounds is denoted by $T$.}
\label{tab:comp}
\resizebox{\textwidth}{!}{%
\begin{tabular}{|c|c|cccc|ccc|}
\hline
\multirow{3}{*}{\textbf{Algorithm}} &
  \multirow{3}{*}{\textbf{Work}} &
  \multicolumn{4}{c|}{\textbf{Characteristics}} &
  \multicolumn{3}{c|}{\textbf{Analysis}} \\ \cline{3-9} 
 &
   &
  \multicolumn{1}{c|}{\multirow{2}{*}{\textbf{Asynchronous}}} &
  \multicolumn{1}{c|}{\textbf{Buffered}} &
  \multicolumn{1}{c|}{\textbf{Quantized}} &
  \textbf{Mitigates} &
  \multicolumn{1}{c|}{\textbf{Unbounded}} &
  \multicolumn{1}{c|}{\textbf{Arbitrary Client}} &
  \textbf{Non-convex objective} \\
 &
   &
  \multicolumn{1}{c|}{} &
  \multicolumn{1}{c|}{\textbf{Aggregation}} &
  \multicolumn{1}{c|}{\textbf{Communication}} &
  \textbf{Error Propagation} &
  \multicolumn{1}{c|}{\textbf{Gradient}} &
  \multicolumn{1}{c|}{\textbf{Update Distribution}} &
  \textbf{$\order{1/\sqrt{T}}$ rate} \\ \hline
\multirow{2}{*}{FedAvg} &
  \cite{communication_efficient} &
  \multicolumn{1}{c|}{\xmark} &
  \multicolumn{1}{c|}{\cmark} &
  \multicolumn{1}{c|}{\xmark} &
  \xmark &
  \multicolumn{1}{c|}{\xmark} &
  \multicolumn{1}{c|}{-} &
  \xmark \\ \cline{2-9} 
 &
  \cite{wang} &
  \multicolumn{1}{c|}{\xmark} &
  \multicolumn{1}{c|}{\cmark} &
  \multicolumn{1}{c|}{\xmark} &
  \xmark &
  \multicolumn{1}{c|}{\cmark} &
  \multicolumn{1}{c|}{-} &
  \cmark \\ \hline
MCM &
  \cite{MCM} &
  \multicolumn{1}{c|}{\xmark} &
  \multicolumn{1}{c|}{\cmark} &
  \multicolumn{1}{c|}{\cmark} &
  \cmark &
  \multicolumn{1}{c|}{\cmark} &
  \multicolumn{1}{c|}{-} &
  \xmark \\ \hline
FedAsync &
  \cite{asyncfl} &
  \multicolumn{1}{c|}{\cmark} &
  \multicolumn{1}{c|}{\xmark} &
  \multicolumn{1}{c|}{\xmark} &
  \xmark &
  \multicolumn{1}{c|}{\xmark} &
  \multicolumn{1}{c|}{\xmark} &
  \cmark \\ \hline
\multirow{2}{*}{FedBuff} &
  \cite{FedBuff} &
  \multicolumn{1}{c|}{\cmark} &
  \multicolumn{1}{c|}{\cmark} &
  \multicolumn{1}{c|}{\xmark} &
  \xmark &
  \multicolumn{1}{c|}{\xmark} &
  \multicolumn{1}{c|}{\xmark} &
  \cmark \\ \cline{2-9} 
 &
  \cite{toghani2022unbounded} &
  \multicolumn{1}{c|}{\cmark} &
  \multicolumn{1}{c|}{\cmark} &
  \multicolumn{1}{c|}{\xmark} &
  \xmark &
  \multicolumn{1}{c|}{\cmark} &
  \multicolumn{1}{c|}{\xmark} &
  \cmark \\ \hline
QAFeL &
  This work &
  \multicolumn{1}{c|}{\cmark} &
  \multicolumn{1}{c|}{\cmark} &
  \multicolumn{1}{c|}{\cmark} &
  \cmark &
  \multicolumn{1}{c|}{\cmark} &
  \multicolumn{1}{c|}{\cmark} &
  \cmark \\ \hline
\end{tabular}%
}
\end{table*}

The main contributions of the manuscript are:
\begin{enumerate}
    \item We introduce QAFeL, an asynchronous FL algorithm with multiple local steps and limited number of communication bits that avoids error propagation using hidden-state updates.
    \item We analytically prove that quantization using QAFeL does not affect the complexity order. More precisely, QAFeL's convergence rate achieves the optimal $\order{1/\sqrt{T}}$ complexity order for non-convex objectives~\cite{complexity} even without assuming uniform client arrival.
    \item We present FedBuff as a special case of QAFeL and fix an error from the original FedBuff manuscript.
          We analytically show that asymptotically the convergence rate of QAFeL is the same as that of FedBuff, without assuming bounded gradients.
    \item We show that the cross-term error caused by staleness and quantization is of smaller order than the errors introduced by each of these factors alone and does not affect the complexity order.
\end{enumerate}
Finally, we validate our theoretical results through experimental evaluation standard benchmarks for FL~\cite{LEAF, CIFAR}.

The rest of the paper is organized as follows.
\cref{sec:system} presents the system model as well as a description of our proposed algorithm.
\cref{sec:algorithm-analysis} presents the analysis of our algorithm and discusses the derived convergence guarantees.
The auxiliary proofs are relegated to Appendix \ref{app:additional-derivations}.
\cref{sec:results-and-discussion} includes experimental results and \cref{sec:conclusion} concludes the manuscript.

\section{System model} \label{sec:system}

In this section, first, we describe our general quantizer model and then present QAFeL in detail.

\subsection{General quantizer model}
A quantizer is composed of an encoder, which receives blocks of information and outputs blocks of bits, and a decoder, which receives blocks of bits and reconstructs blocks of information.
If the encoder and decoder are designed carefully, the mean square error between the original and the reconstructed symbols is small, for example smaller than the norm of the original symbols.
Obviously, the mean square error grows as the compression ratio increases.
Let us denote the combination of the encoder and decoder as a single function $Q$ which is in agreement with the following standard definition of a quantizer in the FL literature \cite{error_feedback, EF21, optimal_compression}.
\begin{definition*}
    \label{def:quantizer}
    A quantizer $Q:\mathbb{R}^d \to \mathbb{R}^d$ (which is a combination of an encoder and a decoder) with a compression parameter $\delta \in (0,1]$ is a (possibly random) function that satisfies
    \begin{equation}
        \mathbb{E}_Q\left[ \normsq{x - Q(x)} \right] \leq (1-\delta) \normsq{x},
    \end{equation}
    where $\mathbb{E}_Q$ is the expectation with respect to the possible internal randomness of the quantizer.
\end{definition*}

\subsection{Proposed algorithm}
\begin{figure}[htbp]
    \centering
    \resizebox{\columnwidth}{!}{%
        \begin{tikzpicture}[
        node distance=1.5cm, auto,
        startstop/.style={rectangle, rounded corners, minimum width=2cm, minimum height=1cm, text centered, draw=black, fill=red!30},
        process/.style={rectangle, rounded corners, minimum width=3cm, minimum height=1cm, text centered, draw=black, fill=cyan!20},
        decision/.style={diamond, aspect=3, minimum width=1cm, minimum height=.5cm, text centered, draw=black, fill=green!20},
        shadedrect/.style={rectangle, rounded corners, minimum width=3cm, minimum height=1cm, text centered, draw=none, dashed, fill=orange!10, fill opacity = 1.00},
        inputoutput/.style={trapezium, trapezium left angle=70, trapezium right angle=110, minimum height=1cm, text centered, draw=black, fill=red!30},
        arrow/.style={thick,->,>=stealth},
        dashedarrow/.style={thick, dashed, ->, >=stealth},
        noarrow/.style={thick, >=stealth}
    ]

    \node [inputoutput] (init) {Initialize $x^0$, $\hat x^0$};
    \node [process, below of=init] (receiveupdate) {Wait for client update};
    \node [process, below of=receiveupdate] (addtobuffer) {Receive update and add to buffer};
    \node [decision, below of=addtobuffer, yshift=-0cm] (checkbuffer) {Is buffer full?};
    \node [process, below of=checkbuffer, yshift=-0.25cm] (globalupdate) {Perform global update};
    \node [decision, below of=globalupdate, yshift=-0.25cm] (checkconvergence) {Reached $T$ updates?};
    \node [inputoutput, below of=checkconvergence, yshift=-.5cm] (stop) {Output $x^t$};
    \node [process, right of=checkconvergence, xshift=4.5cm, text width=4.5cm] (computeqt) {Compute $q^t$, the quantized difference between global model and hidden state};
    \node [process, below of=computeqt] (updatehidden) {Server updates hidden state};
    \node [process, above of=computeqt] (broadcastqt) {Broadcast $q^t$};
    \node [shadedrect, above of=broadcastqt, yshift=.25cm, align=center, minimum width=4.5cm] (clientbackground) {Client background\\~\\~\\~};
    \node [process, above of=broadcastqt] (updatehiddenclient) {Update hidden state};
    \node [shadedrect, above of=updatehiddenclient, yshift=2.25cm, align=center, minimum width=4.5cm] (clientsampled) {Client training\\~\\~\\~\\~\\~\\~\\~\\~\\~\\~};
    \node[process, above of=clientbackground, yshift=0.25cm] (sendupdate) {Send quantized $y^{P}-y^0$};
    \node[process, above of=sendupdate] (localsteps) {Perform $P$ local steps};
    \node[process, above of=localsteps, text width=3.75cm] (copyhiddenstate) {Copy hidden state $y^0 \gets \hat x^t$};

    \draw [arrow] (init) -- (receiveupdate);
    \draw [arrow] (receiveupdate) -- (addtobuffer);
    \draw [arrow] (addtobuffer) -- (checkbuffer);
    \draw [arrow] (checkbuffer.east) -- node[anchor=south] (auxnode) {No}++(1.5,0) |- (receiveupdate.east);
    \draw [arrow] (checkbuffer) -- node[anchor=east] {Yes} (globalupdate);
    \draw [arrow] (globalupdate) -- (checkconvergence);
    \draw [arrow] (checkconvergence) -- node[anchor=east] {Yes} (stop);
    \draw [arrow] (checkconvergence) -- node[anchor=south] {No} (computeqt);
    \draw [arrow] (computeqt) -- (updatehidden);
    \draw [arrow] (computeqt) -- (broadcastqt);
    \draw [dashedarrow] (broadcastqt) -- (updatehiddenclient);
    \draw [noarrow] (broadcastqt) -| ($(auxnode.south) + (.75,0)$);

    \draw [arrow] (copyhiddenstate) -- (localsteps);
    \draw [arrow] (localsteps) -- (sendupdate);
    \draw [dashedarrow] (sendupdate) -- (addtobuffer);

\end{tikzpicture}
    }
    \caption{QAFeL block diagram. Shaded parts occur on the client side. Dashed lines indicate communication between the clients and the server.}
    \label{fig:QAFeL-block-diagram}
\end{figure}
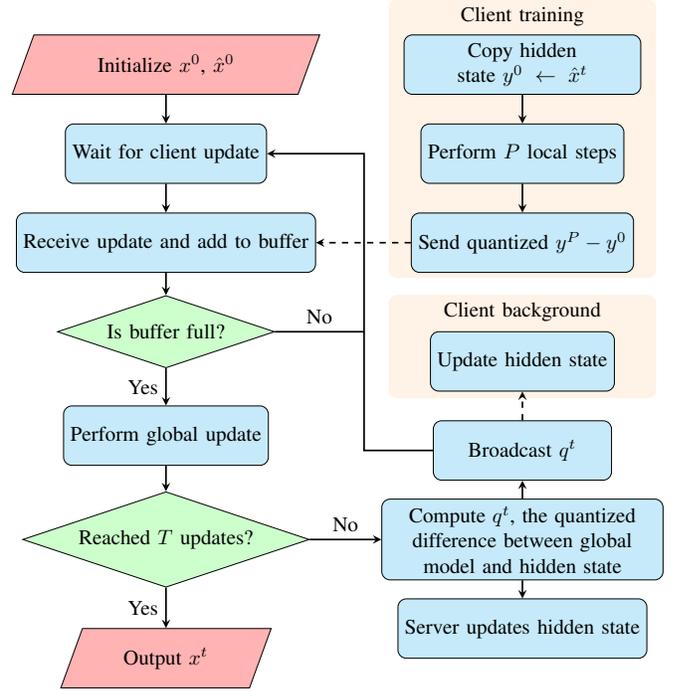

QAFeL is comprised of three processes that run concurrently: QAFeL-server, QAFeL-client and QAFeL-client-background.
The three processes are depicted in \cref{fig:QAFeL-block-diagram} as a block diagram.
The algorithm works as follows.
First, clients initialize the hidden-states with $\hat x^0$, which is the same as the initial server model, $x^0$.
Then, the server waits for a client update.
To start training, the client copies the locally stored hidden-state into a variable $y_0 \gets \hat x^t$, and performs $P$ local model update steps of the type
\begin{equation} \label{eq:local-step}
    y_{p+1} = y_{p} - \eta_\ell \cdot g(y_p),
\end{equation}
where $g(y_p)$ is a noisy, unbiased estimator of the gradient at $y_p$ based on the local dataset and $\eta_\ell$ is a local step-size.
Then, the client sends the quantized update $Q_c(y_0 - y_{P})$ to the server, where $Q_c$ is the client's quantizer.
The server adds the received updates to its buffer.
We denote the $k$-th update in the buffer as $\Delta_k$.
The server keeps receiving updates until its buffer is full, i.e., the server has received $K$ updates.
Then, the server updates the global model by averaging the received updates:
\begin{equation} \label{eq:global_update}
    x^{t+1} = x^t - \eta_g \overline{\Delta}^t, \quad \overline{\Delta}^t\coloneqq \frac{1}{K}\sum_{k=0}^{K-1} \Delta_k.
\end{equation}
When the pre-defined number of iterations $T$ is reached, the server outputs the model and the training stops.
Otherwise, the server computes the quantized difference between its updated model and the hidden-state,
\begin{equation}
    q^t = Q_s(x^{t+1} - \hat x^t),
\end{equation}
where $Q_s$ is the server's quantizer.
The server broadcasts $q^t$ to all clients. Finally, the clients and the server update their copies of the hidden-state using the same equation, $\hat x^{t+1} = \hat x^{t} + q^t$.

For further details on the algorithm design see Appendix \ref{app:pseudo}, where we have included the pseudocode for all components of QAFeL.
The highlighted lines in the pseudocode represent the novel hidden-state mechanism.

\section{Algorithm analysis} \label{sec:algorithm-analysis}
In this section, we analyze the convergence of QAFeL.
First, we formulate the non-convex optimization problem that QAFeL solves.
Then, we present convergence upper bounds for QAFeL.
\subsection{Problem formulation}
We consider the weighted FL problem setting to find a model $x \in \mathbb{R}^d$ such that
\begin{equation} \label{eq:minimization_problem}
    \min_{x \in \mathbb{R}^d} f(x) \coloneqq \sum_{n=1}^N w_n F_n(x),
\end{equation}
where $F_n$ is Client $n$'s loss function, $w_n$ is the client's weight, and $N$ is the total number of clients.
Note that each function $F_n$ implicitly depends on the local data at Client $n$.
The weights $w_n$ can be any arbitrary set of numbers chosen to weigh the importance of each client or to strategically differentiate different clients.
For example, one can assign a larger weight to clients that have a larger dataset.
Obviously, this setting is more general than the equal-weight setting, which is also covered in our setup by choosing $w_n = 1/N$.
Alternatively, one can enforce a uniform request for updates from the server, a technique employed in FedBuff.
While enforcing uniform client participation guarantees that all weights are identical, it slows down the fastest clients' update frequency.
Our analysis considers the case that $f$ is non-convex, which is prevalent in FL applications.

Let us make the following standard assumptions:
\begin{assumption}
    \label{ass:unbiased-stochastic-gradient}
    Unbiased Stochastic Gradients Assumption: We have an unbiased stochastic estimator $g_n(x)$ of the true gradient $\nabla F_n(x)$, i.e., $\expec{g_n(x)} =  \nabla F_n(x)$ for all $x$, at every client.
\end{assumption}
\begin{assumption}
    \label{ass:bounded-local-variance}
    Bounded Local Variance Assumption: At every client,
    \begin{equation}
        \expec{ \normsq{g_n(x) - \nabla F_n(x) } } \leq \sigma_\ell^2, \quad \forall x \in \mathbb{R}^d.
    \end{equation}
\end{assumption}
\begin{assumption}
    \label{ass:L-smooth-and-bounded}
    L-smoothness Assumption: At every client, $F_n$ is $L$-smooth, i.e., $F_n$ is differentiable and its gradient is Lipschitz continuous. Thus, $\nabla F_n$ satisfies
    \begin{equation}
        \norm{\nabla F_n(x) - \nabla F_n(x')} \leq L \norm{x - x'}, \quad \forall x,x'\in \mathbb{R}^d.
    \end{equation}
\end{assumption}
\begin{assumption}
    \label{ass:bounded-heterogeneity}
    Bounded Client Heterogeneity Assumption: The gradients $\nabla F_n$ at every client satisfy
    \begin{equation}
        \normsq{\nabla f(x) - \nabla F_n(x)} \leq B, \quad \forall x\in \mathbb{R}^d.
    \end{equation}
    In the proof, we only use bounded heterogeneity at each buffer set $\calS_t$, i.e., each set of clients that fills the buffer.
    Therefore, we can relax the assumption to
    \begin{equation}
        \frac{1}{K} \sum_{k \in \calS_t }\normsq{\nabla f(x) - \nabla F_k(x)} \leq B, \quad \forall x\in \mathbb{R}^d,
    \end{equation}
    and the results still hold.
    However, for the clearness of exposition, we will use the first version in the proof.
\end{assumption}
\begin{assumption}
    \label{ass:f-lower-bound}
    Lower Bounded Objective Assumption: There exists a lower bound $f^* \leq f(x)$ for all $x \in \mathbb{R}^d$.
\end{assumption}

In an asynchronous setting, different nodes may operate on different versions of the model causing staleness \cite{FedBuff}. The definition of the staleness for QAFeL and a bounded staleness assumption similar to that of FedBuff are presented below.
\begin{definition*}
    For Client $n$, at server step $t$, the difference between the current hidden-state $\hat x^t$ and the hidden-state version that was used to start the current local training is called \emph{staleness}, and is denoted $\tau_n(t)$.
\end{definition*}

\begin{assumption}
    \label{ass:bounded-staleness}
    Bounded Staleness Assumption: At every server step $t$, and for each client $n$, the staleness $\tau_n(t)$ is less than or equal to a maximum allowed staleness, $\tau_{\max, K}$, where $K$ is the server buffer size.
\end{assumption}
Note that as the buffer size increases, the server updates less frequently, which reduces the number of server steps between when a client starts training and when its updates are applied at the server.
If \cref{ass:bounded-staleness} is met for any $K>1$, the maximum delay, $\tau_{\max,K}$, is at most $\lceil \tau_{\max,1}/K \rceil$; this is proven in \cite[Appendix A]{FedBuff}.

In most ML applications, the objective function is highly non-convex \cite{loss-multilayer}.
With such functions, we cannot guarantee the existence of a global minimum.
Instead, as is standard in non-convex optimization, our goal is to find a \emph{first-order $\varepsilon$-stationary point}, that is, $x \in \mathbb{R}^d$ such that $\normsq{\nabla f (x)} \leq \varepsilon$.
With such a goal in mind, we study the ergodic squared norm of the gradient after $T$ iterations:
\begin{equation}
    \frac{1}{T} \sum_{t=0}^{T-1} \expec{\norm{\nabla f(x^t)}^2}.
\end{equation}
We call this quantity the \emph{ergodic convergence rate} of our algorithm.
Upper bounding the rate by $\varepsilon$ ensures that our algorithm finds a first-order $\varepsilon$-stationary point in expectation.

Throughout the paper, we use the notation from \cref{tab:notation-summary}.
\begin{table}[ht]
    \caption{Summary of notation.\label{tab:notation-summary}}
    \centering
    \begin{tabular}{|c||c|}
        \hline
        $x^t, \hat x^t$                   & server, shared hidden-state at time $t$                      \\ \hline
        $L$                               & $L$-smoothness constant of the loss function                 \\ \hline
        $P, p$                            & number, index of local steps at client                       \\ \hline
        $K, k$                            & number, index of clients at the buffer                       \\ \hline
        $N, n$                            & number, index of total clients                               \\ \hline
        $\eta_g, \eta_\ell$               & server, client learning rates                                \\ \hline
        $Q_s, Q_c$                        & server, client quantizers                                    \\ \hline
        $\overline{\Delta}^t, \Delta_k^t$ & server, client $k$'s update at time $t$                      \\ \hline
        $\calS^t$                         & set of client indices at the buffer at time $t$              \\ \hline
        $y_{k,p}^t$                       & local state at client $k$, during local step $p$ at time $t$ \\ \hline
        $\pm$                             & plus and minus a quantity, i.e., $a \pm b = a + b - b = a$   \\
        \hline
    \end{tabular}
\end{table}

\subsection{Convergence analysis for QAFeL}
In this section, we present our main result, as well as a corollary that analyzes the convergence rate in detail.
\begin{theorem} \label{thm:main}
    Consider the optimization problem in \eqref{eq:minimization_problem} satisfying \cref{ass:unbiased-stochastic-gradient,ass:bounded-local-variance,ass:L-smooth-and-bounded,ass:bounded-heterogeneity,ass:bounded-staleness}.
    Then, QAFeL's iterations satisfy
    \begin{align}
        \frac{1}{T} \sum_{t=0}^{T-1} & \expec{\normsq{\nabla f(x^t)}} \leq 4\frac{f(0) - f(x^T)}{T \eta_g P \eta_\ell} \nonumber                              \\
                                     & + 8 L^2  \frac{\eta_g^2P\eta_\ell^2(2-\delta_c) (\tau_{\max, K}^2 + \frac{8}{\delta_s^2}) \sigma_\ell^2}{K} \nonumber \\
                                     & + 80 L^2P^2 \eta_\ell^2 (\sigma_\ell^2 + B) \nonumber                                                                 \\
                                     & + 2L \frac{\eta_g\eta_\ell(2-\delta_c) \sigma_\ell^2 }{K},
    \end{align}
    as long as
    \begin{equation}
        \eta_g^2(\tau_{\max, K}^2 + \frac{8}{\delta_s^2} ) + (1+\frac{1-\delta_s}{K})\eta_\ell \eta_g L \leq \frac{1}{P}
    \end{equation}
    and
    \begin{equation}
        \eta_\ell^2 \leq \frac{1}{80 L^2P^2 \tau_{\max, K}}, \quad \eta_\ell \leq \frac{1}{4L(P+1)}.
    \end{equation}
\end{theorem}
\begin{IEEEproof}
    QAFeL's iterations are
    \begin{align}
        x^{t+1}      & = x^t - \eta_g \overline{\Delta}^t, \label{eq:QAFeL-2} \\
        \hat x^{t+1} & = \hat x^t + Q_s(x^t - \hat x^t).
    \end{align}
    A special case of QAFeL is when the server quantizer $Q_s$ is the identity, for which the hidden-state is the same as the server's state.
    This special case covers FedBuff.
    In other words, our results are applicable to FebBuff as a special case.

    \Cref{ass:L-smooth-and-bounded} implies
    \begin{equation} \label{eq:Taylor-L-smooth-2}
        f(y) \leq f(x) + \langle \nabla f(x), y-x \rangle + \frac{L}{2} \norm{y-x}^2,
    \end{equation}
    for all $x, y \in \mathbb{R}^d$.
    This follows immediately from \cite[Lemma 1.2.3]{Nesterov_2018}.
    Plugging \eqref{eq:QAFeL-2} into \eqref{eq:Taylor-L-smooth-2} yields
    \begin{align}
        f(x^{t+1}) & \leq f(x^t) + \langle \nabla f(x^t), x^{t+1} - x^t \rangle + \frac{L}{2} \normsq{x^{t+1} - x^t}                                                                    \nonumber \\
                   & = f(x^t) - \eta_g \langle \nabla f(x^t), \overline{\Delta}^t \rangle + \eta_g^2\frac{L}{2} \normsq{\overline{\Delta}^t}. \label{eq:Taylor-of-f-2}
    \end{align}
    Using \cref{lemma:expectation-dot-product} and \cref{cor:last-term-cor} from Appendix \ref{app:additional-derivations}, we can add the terms for $t = 0, \ldots, T-1$ and obtain
    \begin{align}
        f(x^{T}) & \leq f(0) - \eta_g \frac{\eta_\ell P}{4} \sum_{t=0}^{T-1} \expec{\normsq{\nabla f(x^t)}} \nonumber                                                               \\
                 & + \eta_g \frac{\eta_\ell P}{2}4L^2 \sum_{t=0}^{T-1} \eta_g^2(\tau_{\max, K}^2 + \frac{8}{\delta_s^2}) \frac{P\eta_\ell^2}{K}(2-\delta_c) \sigma_\ell^2 \nonumber \\
                 & + \eta_g \frac{\eta_\ell P}{2}4L^2 \sum_{t=0}^{T-1} 10P^2 \eta_\ell^2 (\sigma_\ell^2 + B) \nonumber                                                              \\
                 & +\sum_{t=0}^{T-1}  \eta_g^2\frac{L}{2} \frac{P\eta_\ell^2}{K}(2-\delta_c) \sigma_\ell^2, \label{eq:unarranged-main-result}
    \end{align}
    as long as
    \begin{equation}
        P\eta_g^2(\tau_{\max, K}^2 + \frac{8}{\delta_s^2} ) + (1+\frac{1-\delta_s}{K})P\eta_\ell \eta_g L \leq 1
    \end{equation}
    and
    \begin{equation}
        40L^2P^2\eta_\ell^2\tau_{\max, K} \leq \frac{1}{2}, \quad \eta_\ell \leq \frac{1}{4L(P+1)}.
    \end{equation}
    The theorem statement is obtained by re-arranging \eqref{eq:unarranged-main-result}.
\end{IEEEproof}
\begin{corollary}[QAFeL's order of complexity] \label{cor:order-complexity}
    Consider the optimization problem  \eqref{eq:minimization_problem} satisfying \cref{ass:unbiased-stochastic-gradient,ass:bounded-local-variance,ass:L-smooth-and-bounded,ass:bounded-heterogeneity,ass:f-lower-bound,ass:bounded-staleness} and define $F^* \coloneqq f(x^0) - f^*$, where $f^*$ minimizes $f$.
    Choosing $\eta_\ell = \order{K^{-1}P^{-1/2}T^{-1/3}}, \eta_g = \order{KT^{-1/6}}$, and a large enough $T$, QAFeL's iterations satisfy
    \begin{align}
        \frac{1}{T} \sum_{t=0}^{T-1} \expec{\norm{\nabla f(x^t)}^2} & \leq \underbrace{\order{\frac{F^*}{\sqrt{PT}}} + \order{\frac{L^2\sigma_\ell^2(2-\delta_c)}{K\sqrt{PT}}}}_{\text{main error terms}} \nonumber \\
                                                                    & + \underbrace{\order{\frac{L^2P(\sigma_\ell^2 + B)}{K^2 T^{2/3}}}}_{\text{heterogeneity term}} \label{eq:corollary}                           \\
                                                                    & + \underbrace{\order{\frac{L(2-\delta_c) (\tau_{\max, K}^2 + \frac{1}{\delta_s^2}) \sigma_\ell^2}{KT}}}_{\text{staleness term}}. \nonumber
    \end{align}
\end{corollary}
Note that the terms with $(2-\delta_c)$ could have been absorbed by the $\order{\cdot}$ notation, but we have purposefully kept them to highlight the minimal effect of the client quantization.

\cref{cor:order-complexity} yields several insights:
\begin{itemize}
    \item The main error term is $\order{1/\sqrt{T}}$, which is the optimal ergodic convergence rate of SGD for non-convex objectives \cite{complexity}.
    \item The effect of the client quantizer, controlled by $\delta_c$, is relatively small compared to the main error term, since it is divided by the buffer size. Intuitively, we average $K$ client updates at the buffer and it is reasonable that the quantization error order is divided by $K$.
    \item Our theory corroborates the well-established observation in the context of synchronous and unquantized scenarios: increasing the number of local steps, $P$, leads to a faster convergence when the model is far from a good solution, but introduces more drift~\cite{balancing_TCOM}.
          Such a drift is also exacerbated by a higher heterogeneity (higher B).
          This is reflected in the heterogeneity error term, as a larger $P$ is beneficial as long as the main error terms dominate.
          Experiments that showcase the effect of the client drift are presented in \cref{sec:logistic-regression}.
    \item The effect of the server quantizer is a term of order $\order{1/T}$, which is negligible for even a moderate number of server steps $T$.
    \item The error added by staleness, controlled by $\tau_{\max,K}$, and the cross-term error between staleness and quantization are of order $\order{1/T}$, which are also negligible compared to the $\order{1/\sqrt{T}}$ terms.
\end{itemize}
As a special case, taking the limit $\delta_c, \delta_s \to 1$ provides the complexity order of FedBuff, i.e., QAFeL without any quantization scheme, as follows:
\begin{align}
    \frac{1}{T} \sum_{t=0}^{T-1} \expec{\norm{\nabla f(x^t)}^2} & \leq \underbrace{\order{\frac{F^*}{\sqrt{PT}}} + \order{\frac{L^2\sigma_\ell^2}{K\sqrt{PT}}}}_{\text{main error terms}} \nonumber \\
                                                                & + \underbrace{\order{\frac{L^2P(\sigma_\ell^2 + B)}{K^2 T^{2/3}}}}_{\text{heterogeneity term}} \label{eq:fedbuff}                 \\
                                                                & + \underbrace{\order{\frac{L(\tau_{\max, K}^2 + 1) \sigma_\ell^2}{KT}}}_{\text{staleness term}}. \nonumber
\end{align}
Note that our assumptions are weaker than those in FedBuff. The above corrected convergence rate\footnote{The rate that appears in the original AISTATS 2022 paper \cite{FedBuff} has a minor error resulted from inaccuracy in Eq. (20) of that paper.} includes a higher-order $(T^{-2/3})$ heterogeneity term because of our weaker assumptions.

\section{Results and discussion} \label{sec:results-and-discussion}
We present results with two methods of compression.
The first method, $\qsgd$, is an unbiased stochastic quantizer that uses a fixed number of bits per dimension \cite{qsgd}. A higher bit-rate results in a more precise quantizer.
The second method, $\topk$, is biased and only sends the $k$ largest coordinates in absolute value, and does not send the rest \cite{sparsifiedGradientMethods}.
The receiver sets the unsent missing coordinates to zero.
\subsection{Logistic regression experiments} \label{sec:logistic-regression}
To illustrate the fact that QAFeL addresses the error propagation caused by naive direct quantization, we present a set of experiments on a standard logistic regression task and run both algorithms for a comparison.

We consider a logistic regression problem with $\ell_2$ regularization on the \emph{mushrooms} dataset from LIBSVM \cite{LIBSVM}.
The simulation parameters are: 100 clients, delays following a half-normal distribution, server buffer size of 10, client learning rate of 2, server learning rate of 0.1, and $\ell_2$ regularization strength of $1/8124$, where $8124$ is the number of samples in the dataset.

\begin{figure}[htbp]
    \centering
    \subfloat[]{\includegraphics[]{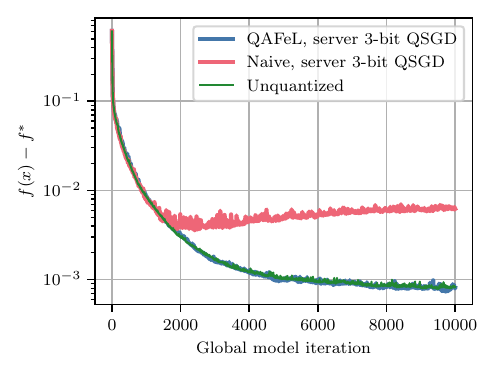}%
        \label{fig:QAFeLvsNaiveQSGD}}
    \\
    \subfloat[]{\includegraphics[]{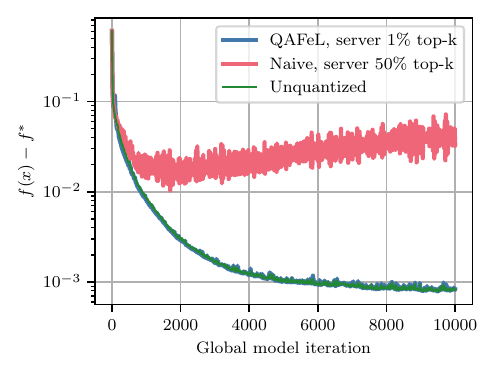}%
        \label{fig:QAFeLvsNaiveTopK}}
    \caption{
        Numerical example of the effect of naive quantization vs. our proposed algorithm (QAFeL).
        No client quantization is performed.
        The $y$-axis illustrates the difference $f(x) - f^\star$, where $f(x)$ is the global model cost at a given iteration, and $f^*$ is the optimal cost.
        Subfigure (a) illustrates the unbiased server quantizer case.
        Subfigure (b) illustrates the biased server quantizer case.}
    \label{fig:QAFeLvsNaive}
\end{figure}

\cref{fig:QAFeLvsNaive} illustrates both the unbiased server quantizer case (\cref{fig:QAFeLvsNaiveQSGD}, $\qsgd$ quantization) and the more drastic biased server quantization case (\cref{fig:QAFeLvsNaiveTopK}, $\topk$ quantization).
In \cref{fig:QAFeLvsNaiveQSGD}, a quantizer with 3 bits is used for both direct quantization and QAFeL. While QAFeL provides results very close to that of the unquantized case, the direct quantization approach does not converge.
Similarly, in the case of $\topk$ quantization in \cref{fig:QAFeLvsNaiveTopK}, QAFeL elegantly manages the error propagation while a direct quantization suffers from error propagation. More specifically, with direct quantization, the algorithm diverges when the server sends the top 50\% of coordinates in absolute value.
Meanwhile, with QAFeL, a server that sends only the top 1\% of coordinates in absolute value converges.

\begin{figure}[htbp]
    \centering
    \includegraphics[]{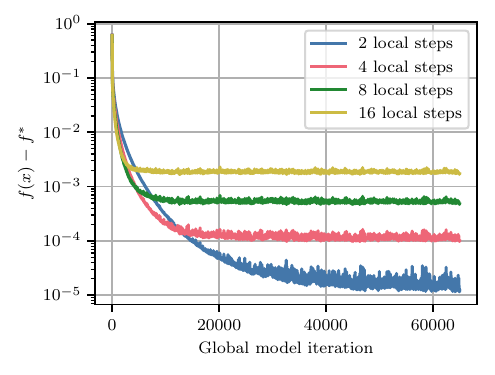}
    \caption{Numerical example of the effect of local steps. The $y$-axis illustrates the difference $f(x) - f^\star$, where $f(x)$ is the global model cost at a given iteration, and $f^*$ is the optimal cost.} QAFeL is run without quantization and a varying quantity of local steps. More local steps lead to larger client drift and faster convergence, but to a less optimal solution.
    \label{fig:logistic-regression}
\end{figure}

\cref{fig:logistic-regression} presents a numerical illustration of the effect of the local step values, without quantization.
For an analogous figure in the synchronous and unquantized case, see \cite[Figure 6]{tighter_theory}.
Observe that a larger amount of local steps implies faster convergence, but higher suboptimality.
The noise around the optimum is due to the randomness introduced by the delays.
Also, note that the noise does not increase for smaller local steps.
Such an appearance is because of the y-axis logarithmic scale.

\subsection{Neural Network experiments}
\subsubsection{Experiment setup}
We have implemented QAFeL using FL Simulator (FLSim) \cite{flsim}, a library written in PyTorch \cite{pytorch}.
To verify our theoretical results, we consider two image classification tasks using the CIFAR-10 \cite{CIFAR} and CelebA \cite{celeba} datasets, as well as a next-word prediction task using the Shakespeare dataset \cite{communication_efficient}.

The CIFAR-10 dataset consists of labeled images pertaining to one of 10 classes.
As is standard in image classification, we first normalize the images using the dataset mean and standard deviation such that each color channel has mean 0 and standard deviation 1.
To simulate a FL scenario, we follow the approach of \cite{dirichlet} to synthesize non-identical clients.
Specifically, we split the dataset into 5,000 clients where each client has a number of samples that follows a symmetric Dirichlet distribution with parameter 0.1.
This parameter controls the correlation of the data distribution, i.e., how close the distribution is to an i.i.d. random process.
When the parameter is $0$, each client holds data from only one class chosen at random.
On the other hand, when it tends to $\infty$, all clients have identical distributions.

The CelebA dataset contains labeled images of celebrities.
We detect whether they are smiling or not.
We resize and crop the images to $32\times 32$ pixels before normalizing them to have 0.5 mean and 0.5 standard deviation, as done in previous work~\cite{FedBuff}.
We also use the default data-heterogeneous client partition from LEAF.

The Shakespeare dataset contains Shakespeare's plays, separated by play and character. This was proposed as a good FL scenario in \cite{communication_efficient} as different characters have different word distributions. 
We use LEAF's default client partition for this task as well.

For both image datasets, we train the image classification model used in LEAF, that is, a four-layer convolutional neural network (CNN) classifier, slightly modified by replacing batch normalization layers with group normalization layers \cite{groupnorm} following the approach of \cite{batchnorm_with_groupnorm}.
We use a 0.1 dropout regularization rate, stride of 1, and padding of 2.
We compare FedBuff and QAFeL with the hyperparameter selection from \cite{FedBuff}, which we re-state in \Cref{tab:hyper} for completeness.
Note that we have used server momentum $\beta$ in two of our experiment settings, which we have found to help convergence in practice.

In our image experiments, we simulate QAFeL training until we reach a pre-specified target evaluation accuracy, which is 60\% for CIFAR-10 and 90\% for CelebA.
These quantities are the same as previous benchmarks \cite{FedBuff, LEAF}, i.e., the pre-defined architectures from our LEAF benchmark converge to these percentages on the centralized setting.

For the Shakespeare dataset, we use the 2-layer stacked LSTM model proposed in \cite{communication_efficient,LEAF}, as well as their proposed target accuracy of 54\%.
We use Bayesian optimization~\cite{bayesian} to obtain the hyperparameters reported in \Cref{tab:hyper}.
\begin{table}[htbp]
    \centering
	\caption{Chosen hyperparameters}
	\label{tab:hyper}
	\begin{tabular}{llllll} 
		\toprule
		         & FedBuff and QAFeL                        & FedAsync                                    \\
		\hline
		
		         & $\eta_{\ell}=4.7\cdot10^{-6}$  & $\eta_{\ell}=5.7$            \\     
		CelebA   & $\eta_g=1.0\cdot10^{3}$        & $\eta_g=2.8\cdot10^{-3}$          \\ 
		         & $\beta=3.0\cdot10^{-1}$        &      $\beta=0$                              \\ 
		\hline
		         & $\eta_{\ell}=1.95\cdot10^{-4}$ & $\eta_{\ell}=1.0\cdot10^{2}$   \\     
		CIFAR-10 & $\eta_g=4.09\cdot10^1$         & $\eta_g=6.4\cdot10^{-5}$         \\ 
		         & $\beta=0$                      &    $\beta=0$                        \\
        \hline
		         & $\eta_{\ell}=1.17$     & $\eta_{\ell}=1.7\cdot10^1$   \\     
		Shakespeare  & $\eta_g=2.64\cdot10^{-1}$       & $\eta_g=2.03\cdot10^{-1}$                  \\ 
		         & $\beta=0$        &         $\beta=5.0\cdot10^{-1}$                              \\ 
		\bottomrule
	\end{tabular}
\end{table}

We simulate client arrival times and training durations based on Meta's production FL system \cite[Appendix C]{FedBuff}.
In this model, clients arrive at a constant rate and their training durations are sampled from a half-normal distribution denoted as $Y$, where $Y = |X|$ and $X \sim \mathcal{N} (0, 1)$.
To achieve varying concurrencies (the maximum number of users training in parallel) of 100, 500, and 1,000 users, we adjust the arrival rates of clients to 125, 627, and 1,253 clients per unit of time, respectively.
These rates are determined based on the expected value $\sqrt{2/\pi}$ of the half-normal distribution $Y$.
As done in previous work \cite{asyncfl, FedBuff}, we use weight scaling to penalize staleness.
More precisely, an update with staleness $\tau$ is scaled down by a $1/\sqrt{1 + \tau}$ factor.
For all scenarios, we use a batch size of 32 and a buffer size $K=10$.

To justify our choice of 4-bit $\qsgd$ quantizer for server and client in our concurrency experiments, we perform a set of experiments displayed in \cref{tab:qsgd_table}, where we assume a constant client arrival rate of 100 clients per unit of time, without performing any weight scaling.
These experiments show that using a 4-bit $\qsgd$ at both client and server does not significantly alter the number of client uploads to reach a target accuracy, yet approximately provides an eight-fold reduction in the number of communicated bits. 

\subsubsection{Experimental results}
We conduct experiments three times and provide the mean and standard deviation of the results.
Alongside the conventional metric of comparing synchronous and asynchronous FL methods, which is the number of client trips, we also report the number of bytes sent per message to emphasize QAFeL's advantages.
\begin{figure}[htbp]
    \centering
    \subfloat[]{\includegraphics[width=.72\columnwidth]{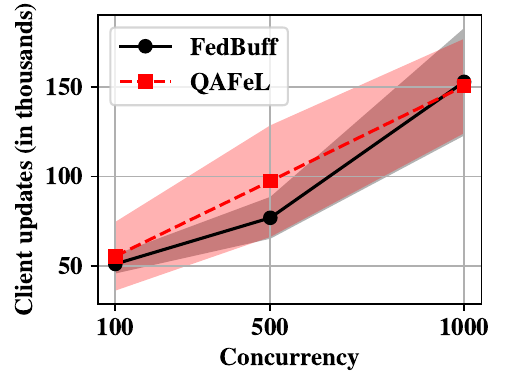}%
        \label{tripsCIFAR}}
    \\
    \subfloat[]{\includegraphics[width=.72\columnwidth]{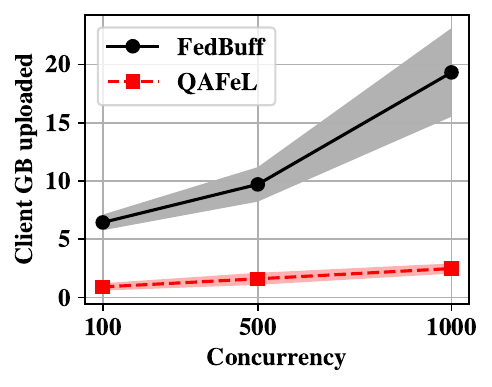}%
        \label{mbsCIFAR}}
    \\
    \subfloat[]{\includegraphics[width=.72\columnwidth]{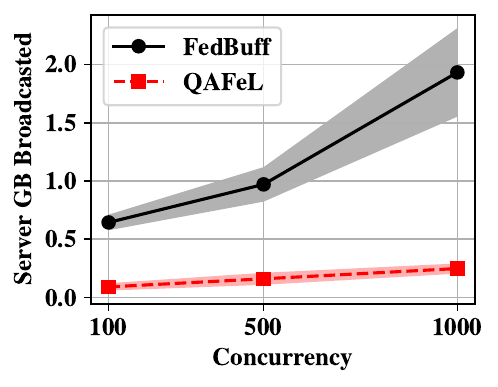}%
        \label{mbsBroadcastedCIFAR}}
    \caption{Comparison of communication metrics for QAFeL and FedBuff across varying concurrency levels (clients training in parallel) to reach 60\% target accuracy on the CIFAR-10 dataset. QAFeL employs 4-bit $\qsgd$ quantization at both the server and client: (a) the number of client updates in thousands, (b)   the total GB uploaded by the clients, and (c) the GB broadcasted by the server.}
    \label{fig:QAFeLvsFedBuff-CIFAR}
\end{figure}
\begin{figure}[htbp]
    \centering
    \subfloat[]{\includegraphics[width=.72\columnwidth]{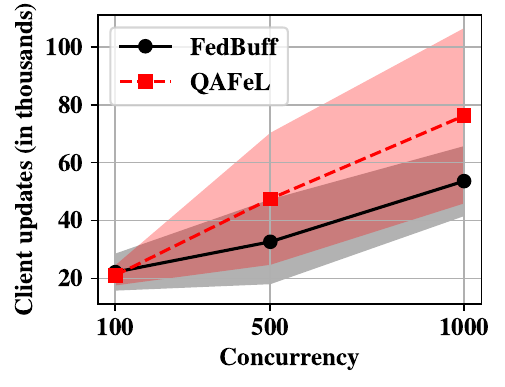}%
        \label{tripsCelebA}}
    \\
    \subfloat[]{\includegraphics[width=.72\columnwidth]{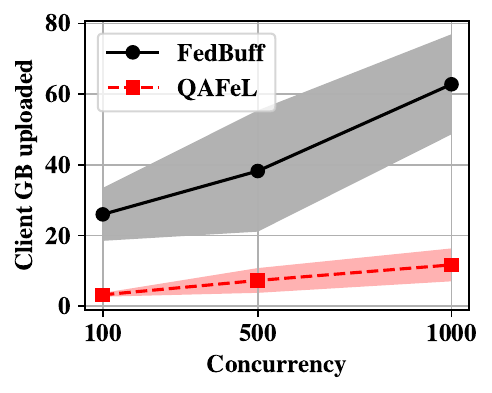}%
        \label{mbsCelebA}}
    \\
    \subfloat[]{\includegraphics[width=.72\columnwidth]{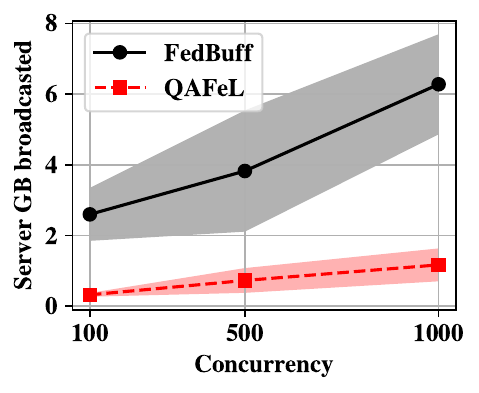}%
        \label{mbsBroadcastedCelebA}}
    \caption{Comparison of communication metrics for QAFeL and FedBuff across varying concurrency levels (clients training in parallel) to reach 90\% target accuracy on the CelebA dataset. QAFeL employs 4-bit $\qsgd$ quantization at both the server and client: (a) the number of client updates in thousands, (b) the total GB uploaded by the clients, and (c) the GB broadcasted by the server.}
    \label{fig:QAFeLvsFedBuff-CelebA}
\end{figure}
\begin{figure}[htbp]
    \centering
    \subfloat[]{\includegraphics[width=.72\columnwidth]{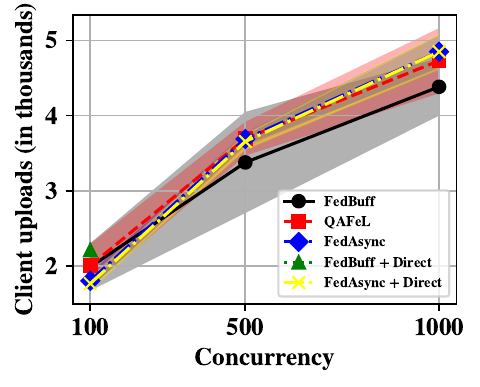}%
        \label{tripsShake}}
    \\
    \subfloat[]{\includegraphics[width=.72\columnwidth]{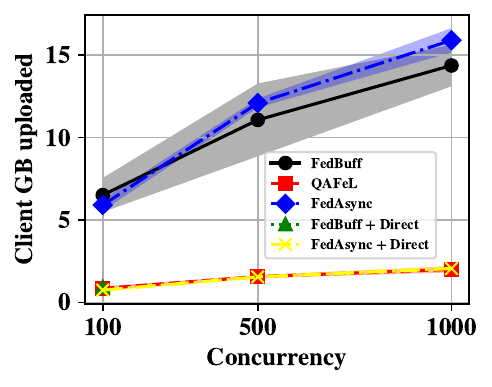}%
        \label{mbsShake}}
    \\
    \subfloat[]{\includegraphics[width=.72\columnwidth]{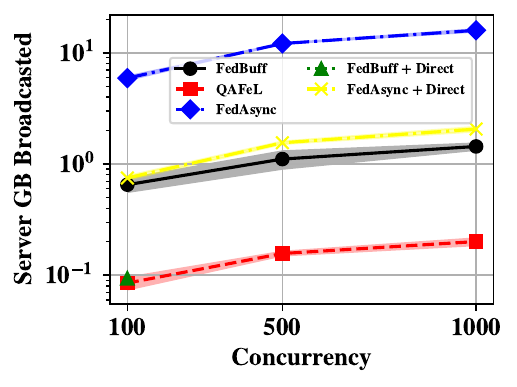}%
        \label{mbsBroadcastedShake}}
    \caption{Comparison of communication metrics for QAFeL, FedBuff, FedAsync, FedBuff  with Direct Quantization, and FedAsync with Direct Quantization across varying concurrency levels (clients training in parallel) to reach 54\% target accuracy on the Shakespeare dataset. QAFeL employs 4-bit $\qsgd$ quantization at both the server and client: (a) the number of client updates in thousands, (b) the total GB uploaded by the clients, and (c) the GB broadcasted by the server. FedBuff with Direct Quantization only has one data-point because for concurrencies 500 and 1000 the algorithm diverged for at least one of the three seeds.}
    \label{fig:QAFeLvsFedBuff-Shake}
\end{figure}

\cref{fig:QAFeLvsFedBuff-CIFAR,fig:QAFeLvsFedBuff-CelebA} illustrate the benefits of QAFeL for various concurrency values over CIFAR-10 and CelebA datasets, respectively.
As shown in Figs. \ref{mbsCIFAR} and \ref{mbsCelebA}, for all concurrencies, QAFeL clients use less uploaded bytes compared to FedBuff.
For example, reduction factors of 5.2 to 8 for the CelebA dataset and 6 to 7.7 for the CIFAR-10 dataset are observed.
Similar reductions for the number of broadcasted bytes by the server are seen in Figs. \ref{mbsBroadcastedCIFAR} and \ref{mbsBroadcastedCelebA}.
Note that the total number of bytes includes the extra client updates that occur by adding quantization.
Notably, the number of client updates remains similar for QAFeL and FedBuff for both datasets.
For example, the maximum increase is about 1.5 times in the CelebA case in \cref{tripsCelebA}.
For both tasks, a 4-bit $\qsgd$ quantizer \cite{qsgd} was used in both upload and download.

\begin{figure}[htbp]
    \centering
    \includegraphics[width=0.72\columnwidth]{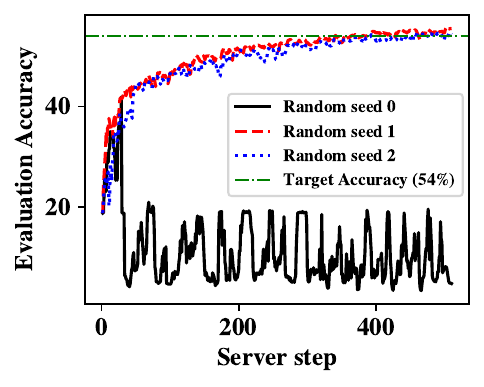}
    \caption{Evaluation curves for the Shakespeare experiments with concurrency 500 and FedBuff with direct 4-bit quantization for both client and server. We observe that one of the three runs diverges.}
    \label{fig:seedsShakespeare}
\end{figure}

\cref{fig:QAFeLvsFedBuff-Shake} illustrates similar results by comparing QAFeL with FedBuff, FedAsync~\cite{asyncfl}, and FedBuff with direct quantization (4-bit $\qsgd$ for both client and server quantizers). 
The latter is not plotted for concurrencies 500 and 1000 as it diverged for at least one of the three experiments that we average over. \cref{fig:seedsShakespeare} shows the results for FedBuff with direct quantization at concurrency 500. The figure clearly shows that FedBuff with direct quantization does not converge for the random seed 0. 
This is consistent with the theory and the results shown in the logistic regression experiments of \cref{fig:FedBuffvsNaive}, which also showed a divergence for the direct quantization case.
Moreover, \cref{fig:QAFeLvsFedBuff-Shake} also corroborates the results shown in \cref{fig:QAFeLvsFedBuff-CIFAR,fig:QAFeLvsFedBuff-CelebA}, as QAFeL performs markedly better than existing algorithms in terms of both uploaded and broadcasted bytes and maintains approximately the same amount of client uploads.
Note that FedAsync is not massively scalable as it does not have a buffer to accumulate client updates, which is reflected in the number of broadcasted bytes.
With the three random seeds that we picked, FedAsync with direct quantization performs close to the unquantized FedAsync in number of client uploads.
However, note that using FedAsync, the number of server broadcasts is  ten times more than those of FedBuff and QAFeL, due to the lack of buffer. 
Also, even with quantization, FedAsync still broadcasts more bytes compared to FedBuff without quantization and an order of magnitude more than QAFeL.
If one were to combine a hidden-state scheme with FedAsync, this would require more computations at the client side, as they would update their hidden states more often due to the high frequency of server updates.

Results for a simple scenario with client arrival rate of 100 and different number of quantization bits are presented in \cref{tab:qsgd_table}.
\begin{table}[htbp]
    \caption{Communication metrics of QAFeL to reach the CelebA target validation accuracy (90\%) with different $\qsgd$ combinations.}
    \label{tab:qsgd_table}
    \centering
    \resizebox{\columnwidth}{!}{%
        \begin{tabular}{|cc|c|c|c|}
            \hline
            \multicolumn{2}{|c|}{\textbf{Algorithm}} & \multirow{2}{*}{\textbf{\begin{tabular}[c]{@{}c@{}}Uploads \\ (thousands)\end{tabular}}} & \multirow{2}{*}{\textbf{kB/upload}} & \multirow{2}{*}{\textbf{kB/download}}          \\ \cline{1-2}
            \multicolumn{1}{|c|}{}                   & \begin{tabular}[c]{@{}c@{}}(client, server) \\ $\qsgd$ bits\end{tabular}                 &                                     &                                       &        \\ \cline{2-5}
            \multicolumn{1}{|c|}{}                   & (8,8)                                                                                    & $27.6 \pm 4.82$                     & 29.924                                & 29.924 \\ \cline{2-5}
            \multicolumn{1}{|c|}{}                   & (8,4)                                                                                    & $29.8 \pm 2.76$                     & 29.924                                & 15.380 \\ \cline{2-5}
            \multicolumn{1}{|c|}{}                   & (8,2)                                                                                    & $35.7 \pm 14.5$                     & 29.924                                & 8.108  \\ \cline{2-5}
            \multicolumn{1}{|c|}{QAFeL}              & (4,8)                                                                                    & $39.7 \pm 3.32$                     & 15.380                                & 29.924 \\ \cline{2-5}
            \multicolumn{1}{|c|}{}                   & (4,4)                                                                                    & $27.8 \pm 10.4$                     & 15.380                                & 15.380 \\ \cline{2-5}
            \multicolumn{1}{|c|}{}                   & (4,2)                                                                                    & $37.7 \pm 9.10$                     & 15.380                                & 8.108  \\ \cline{2-5}
            \multicolumn{1}{|c|}{}                   & (2,8)                                                                                    & $58.6 \pm 7.16$                     & 8.108                                 & 29.924 \\ \cline{2-5}
            \multicolumn{1}{|c|}{}                   & (2,4)                                                                                    & $74.6 \pm 35.7$                     & 8.108                                 & 15.380 \\ \cline{2-5}
            \multicolumn{1}{|c|}{}                   & (2,2)                                                                                    & $91.9 \pm 34.4$                     & 8.108                                 & 8.108  \\ \hline
            \multicolumn{2}{|c|}{FedBuff}            & $26.1 \pm 6.7$                                                                           & 117.128                             & 117.128                                        \\ \hline
        \end{tabular}%
    }
\end{table}
Using fewer quantization bits at the server always leads to fewer total downloaded bytes.
However, using less precision at the client-side, for example reducing the number of quantization bits from 4 to 2 while keeping the server's quantizer at 2 bits per sample, may result in more total uploaded bytes.
Moreover, \cref{tab:qsgd_table} shows an increase in the number of uploads by switching from 4 to 2 bits per sample at the client-side resulting in similar total upload bytes.
This situation highlights a trade-off between the level of quantization and how quickly the system converges.

In simpler terms, while reducing the compression ratio means transmitting fewer bytes, it may require more messages to achieve the desired accuracy.
Therefore, it is important to find the right trade-off between the two factors.
Nevertheless, QAFeL generally uses fewer total communication bits compared to FedBuff.
Also, our simulation results indicate that the best compression ratio depends on the chosen quantization method.

\section{Conclusion} \label{sec:conclusion}
We propose a practical, scalable, and communication-efficient FL algorithm that allows client asynchrony and quantized communications.
Using a hidden-state scheme, we avoid error propagation.
We provide optimal ergodic convergence rates for SGD on non-convex objectives, without assuming bounded gradients or uniform client arrival.
We also show that the cross-term error between staleness and quantization is negligible compared to each of the error terms.
These are achieved without assuming uniform client arrival or imposing restrictions on step-size choices.
The presented theoretical analysis can guide the design of quantizers to achieve a specific convergence rate and has been corroborated empirically.
A common application is the design of FL systems with bandwidth constraints.
Our simulation results show that the effects of the client quantization are larger than those of the server quantization.

\appendices
\section{Additional derivations} \label{app:additional-derivations}
This appendix contains the derivations of the necessary lemmas for the main result.
First, we present some useful remarks:
\begin{remark} \label{remark:sum-unbiasedly-quantized-bound}
    Given an unbiased quantizer $Q$ with parameter $\delta$, for any set of $N \geq 1$ vectors $\{x_n \in \mathbb{R}^d, n = 1,\ldots, N\}$, we have
    \begin{equation*}
        \expecq {\norm{\sum_{n=1}^{N} Q(x_n)}^2} \leq \norm{\sum_{n = 1}^N x_n}^2 + (1-\delta) \sum_{n=1}^N \norm{x_n}^2.
    \end{equation*}
\end{remark}
The remark follows immediately from the unbiasedness of $Q$ and our definition of a quantizer, see \cref{def:quantizer}.

\begin{lemma} \label{lemma:expectation-dot-product}
    Consider the optimization problem in \eqref{eq:minimization_problem} satisfying \cref{ass:unbiased-stochastic-gradient,ass:bounded-local-variance,ass:L-smooth-and-bounded,ass:bounded-staleness,ass:bounded-heterogeneity}.
    Then, QAFeL's iterations satisfy
    \begin{multline}
        \expec{-\langle \nabla f(x^t), \overline{\Delta}^t \rangle} \leq -\frac{P\eta_\ell}{2} \expec{\normsq{\nabla f(x^t)}} - \frac{\eta_\ell}{2} r_t \\
        +  P \eta_\ell L^2  \mathbb{E} \biggr [ 2\tau_{\max, K} \eta_g^2 \sum_{s = t - \tau_{\max, K}}^{t-1} \normsq{\overline{\Delta}^s} \\
        + \frac{12}{\delta_s} \eta_g^2 \sum_{s=1}^{t-1}(1- \frac{\delta_s}{2})^s \normsq{\overline{\Delta}^{t- s - 1}} \\
        + 20P^2 \eta_\ell^2 (\sigma_\ell^2 +  B + \sum_{s=1}^{\tau_{\max, K}}\normsq{\nabla f(x^{t-s})})\biggr ],
    \end{multline}
    where $r_t = \sum_{p=0}^{P-1} \expec{\frac{1}{K} \sum_{k\in \calS_t} \normsq{ \nabla F_k(y^{t-\tau_k}_{k,p}) }}$.
\end{lemma}
\begin{IEEEproof}
    Let us start by expanding the definition of the server update, which is the average of client updates,
    \begin{equation} \label{eq:server-update-definition-2}
        \overline{\Delta}^t = \frac{1}{K} \sum_{k\in \calS_t} \Delta^{t-\tau_k}_k.
    \end{equation}
    We have defined the set of clients that participate in the update at time $t$ as $\calS_t$ and the update from Client $k$ as $\Delta^{t-\tau_k}_k$, where $t-\tau_k$ represents the time index of the hidden-state used for the update.
    Also, $\Delta^{t-\tau_k}_k = Q_c(y^{t-\tau_k}_{k,0} - y^{t-\tau_k}_{k,P} )$, where $y^{t-\tau_k}_{k,p}$ is the local model after $p$ local updates.
    The initial model used for updating $\Delta^{t-\tau_k}_k$ is $y^{t-\tau_k}_{k,0} = \hat x^{t-\tau_k}$.
    Now, since the client quantizer $Q_c$ is unbiased, we have
    \begin{equation}
        \begin{aligned}
            \expec{\Delta^{t-\tau_k}_k} & = \expec{y^{t-\tau_k}_{k,0} - y^{t-\tau_k}_{k,P}}                    \\
                                        & = \expec{\sum_{p=0}^{P-1} \eta_\ell g_k(y^{t-\tau_k}_{k,p})}         \\
                                        & = \expec{\sum_{p=0}^{P-1} \eta_\ell \nabla F_k(y^{t-\tau_k}_{k,p})},
        \end{aligned}
    \end{equation}
    where the last equality is derived from \cref{ass:unbiased-stochastic-gradient}.
    Then, applying the linearity of the expectation and using the last two equations show that $\expec{\langle \nabla f(x^t), \overline{\Delta}^t \rangle}$ is
    \begin{equation}
        \expec{\left \langle \nabla f(x^t), \frac{1}{K} \sum_{k\in \calS_t} \sum_{p=0}^{P-1} \eta_\ell \nabla F_k(y^{t-\tau_k}_{k,p}) \right \rangle}.
    \end{equation}
    We re-arrange and obtain
    \begin{equation}
        \sum_{p=0}^{P-1} \eta_\ell \expec{  \frac{1}{K} \sum_{k\in \calS_t}\left \langle \nabla f(x^t),  \nabla F_k(y^{t-\tau_k}_{k,p}) \right \rangle}.
    \end{equation}
    Using the well-known identity $\langle a, b \rangle =  \tfrac{1}{2} (\normsq{a} + \normsq{b} - \normsq{a - b})$, and the definition of $r_t$, we obtain
    \begin{multline} \label{eq:negative_term}
        \expec{-\langle \nabla f(x^t), \overline{\Delta}^t \rangle} = -\frac{P\eta_\ell}{2} \expec{\normsq{\nabla f(x^t)}} - \frac{\eta_\ell}{2} r_t \\
        {} +  \sum_{p=0}^{P-1} \frac{\eta_\ell}{2}  \expec{\frac{1}{K} \sum_{k\in \calS_t} \normsq{ \nabla f(x^t) - \nabla F_k(y^{t-\tau_k}_{k,p}) }}.
    \end{multline}
    For the last term, we can add and subtract $\nabla f(y^{t-\tau_k}_{k,p})$, apply Cauchy-Schwarz and use \cref{ass:L-smooth-and-bounded} to obtain
    \begin{equation} \label{eq:two-lipschitz}
        \mathbb{E} \normsq{ \nabla f(x^t) - \nabla F_k(y^{t-\tau_k}_{k,p}) } \leq 2L^2 \mathbb{E} \normsq{x^t - y^{t-\tau_k}_{k,p} }.
    \end{equation}
    Applying Cauchy-Schwarz once again, we obtain a decomposition in two error terms
    \begin{equation*}
        \normsq{x^t - y^{t-\tau_k}_{k,p} } \leq 2 \underbrace{\normsq{x^t - x^{t-\tau_k}}}_{\text{staleness}} + 2\underbrace{\normsq{x^{t-\tau_k} - y^{t-\tau_k}_{k,p}}}_{\text{local drift}}.
    \end{equation*}
    We bound the staleness term with \cref{ass:bounded-staleness},
    
    \begin{equation}
        \begin{aligned}
            \normsq{x^t - x^{t-\tau_k}} & \leq \tau_{\max, K} \sum_{s = t - \tau_{\max, K}}^{t-1} \normsq{x^{s+1} - x^{s}}            \\
                                        & = \tau_{\max, K} \eta_g^2 \sum_{s = t - \tau_{\max, K}}^{t-1} \normsq{\overline{\Delta}^s}.
        \end{aligned}
    \end{equation}
    Using \cref{lemma:local-updates-bound} with $z = x^{t-\tau_k}$ we bound the local drift term with
    \begin{multline}
        \expec{\normsq{y^{t-\tau_k}_{k,p} - y^{t-\tau_k}_{k,0}}} \leq 10P^2 \eta_\ell^2 (\sigma_\ell^2 + \expec{\normsq{\nabla f(x^{t-\tau_k})}} \\
        {} + \expec{\normsq{\nabla F_k(x^{t-\tau_k}) - \nabla f(x^{t-\tau_k})}}) \\
        {} + 3 \underbrace{\normsq{\hat x^{t - \tau_k} - x^{t-\tau_k}}}_{\text{server quantization}}.
    \end{multline}
    Then, using \cref{ass:bounded-staleness,ass:bounded-heterogeneity}, we obtain
    \begin{multline*}
        \expec{\normsq{y^{t-\tau_k}_{k,p} - y^{t-\tau_k}_{k,0}}} \leq 10P^2 \eta_\ell^2 (\sigma_\ell^2 +  B) \\
        {} + 10P^2 \eta_\ell^2 \sum_{s=1}^{\tau_{\max, K}} \expec{\normsq{\nabla f(x^{t-s})}} \\
        {}+ 3 \underbrace{\normsq{\hat x^{t - \tau_k} - x^{t-\tau_k}}}_{\text{server quantization}}.
    \end{multline*}
    We can bound the server quantization term as
    \begin{multline}
        \mathbb{E}_Q \normsq{x^{t-\tau_k} - \hat x^{t-\tau_k}} \leq (1+c^{-1}) \eta_g^2 \normsq{\overline{\Delta}^{t-\tau_k - 1}}  \\
        {} +(1+c)(1-\delta_s) \normsq{x^{t-\tau_k - 1} - \hat x^{t-\tau_k - 1}}.
    \end{multline}
    Selecting $c = \delta_s/2$, we obtain
    \begin{multline}
        \mathbb{E}_Q \normsq{x^{t-\tau_k} - \hat x^{t-\tau_k}} \leq \frac{2}{\delta_s} \eta_g^2 \normsq{\overline{\Delta}^{t-\tau_k - 1}} \\
        {} + (1- \frac{\delta_s}{2}) \normsq{x^{t-\tau_k - 1} - \hat x^{t-\tau_k - 1}}.
    \end{multline}
    Inductively, this yields
    \begin{equation*}
        \mathbb{E}_Q \normsq{x^{t-\tau_k} - \hat x^{t-\tau_k}} \leq  \frac{2}{\delta_s} \eta_g^2 \sum_{s=1}^{t-1}(1- \frac{\delta_s}{2})^s \normsq{\overline{\Delta}^{t- s - 1}}.
    \end{equation*}
    Substituting the staleness, server quantization, and local drift terms provides
    \begin{align*}
        \expec{\normsq{x^t - y^{t-\tau_k}_{k,p} }} & \leq 2 \tau_{\max, K} \eta_g^2 \sum_{s = t - \tau_{\max, K}}^{t-1} \expec{\normsq{\overline{\Delta}^s}}               \\
                                                   & + \frac{12}{\delta_s} \eta_g^2 \sum_{s=1}^{t-1}(1- \frac{\delta_s}{2})^s \expec{\normsq{\overline{\Delta}^{t- s - 1}}} \\
                                                   & + 20P^2 \eta_\ell^2 (\sigma_\ell^2 +  B)                                                                              \\
                                                   & + 20P^2 \eta_\ell^2\sum_{s=1}^{\tau_{\max, K}} \expec{\normsq{\nabla f(x^{t-s})}}.
    \end{align*}
    Substituting in \eqref{eq:two-lipschitz} and then \eqref{eq:negative_term} proves the lemma.
\end{IEEEproof}

\begin{lemma} \label{lemma:expectation-of-last-term}
    Consider the optimization problem in \eqref{eq:minimization_problem} satisfying \cref{ass:unbiased-stochastic-gradient,ass:bounded-local-variance}.
    Then, QAFeL's iterations satisfy
    \begin{multline}
        \expec{ \norm{ \overline{\Delta}^t }^2 } \leq \frac{P\eta_\ell^2}{K}(2-\delta_c) \sigma_\ell^2 \\
        {} + (1 + \frac{1-\delta_c}{K}) \expec{ \frac{1}{K}  \sum_{k\in \calS_t}  \norm{\sum_{p=0}^{P-1} \eta_\ell \nabla F_k(y^{t-\tau_k}_{k,p}) }^2 }.
    \end{multline}
\end{lemma}
\begin{IEEEproof}
    First, using the unbiasedness of the client quantizer, we apply \cref{remark:sum-unbiasedly-quantized-bound} as follows
    \begin{equation*}
        \begin{aligned}
            \expec{ \norm{ \overline{\Delta}^t }^2 }
             & = \expec{  \norm{\frac{1}{K}\sum_{k\in \calS_t} Q_c \left( \sum_{p=0}^{P-1} \eta_\ell g_k(y^{t-\tau_k}_{k,p}) \right) }^2 } \\
             & \leq \expec{\norm{\frac{1}{K}\sum_{k\in \calS_t} \sum_{p=0}^{P-1} \eta_\ell g_k(y^{t-\tau_k}_{k,p})}^2}                     \\
             & + (1-\delta_c) \expec{ \frac{1}{K^2}\sum_{k\in \calS_t}  \norm{\sum_{p=0}^{P-1} \eta_\ell g_k(y^{t-\tau_k}_{k,p}) }^2 }.
        \end{aligned}
    \end{equation*}
    Then, using the unbiasedness of the stochastic gradient on both terms completes the proof.
\end{IEEEproof}

\begin{corollary} \label{cor:last-term-cor}
    Consider the optimization problem in \eqref{eq:minimization_problem} satisfying \cref{ass:unbiased-stochastic-gradient,ass:bounded-local-variance}.
    Then, QAFeL's iterations satisfy
    \begin{equation}
        \expec{ \norm{ \overline{\Delta}^t }^2 } \leq \frac{P\eta_\ell^2}{K}(2-\delta_c) \sigma_\ell^2 + (1 + \frac{1-\delta_c}{K}) P \eta_\ell^2 r_t,
    \end{equation}
    where $r_t = \sum_{p=0}^{P-1} \expec{ \frac{1}{K}  \sum_{k\in \calS_t}  \norm{ \nabla F_k(y^{t-\tau_k}_{k,p}) }^2 } $.
\end{corollary}

The following Lemma is a generalization of \cite[Equations (52) to (64)]{toghani2022unbounded}.
\begin{lemma} \label{lemma:local-updates-bound}
    Given any $z \in \mathbb{R}^d$, if \cref{ass:bounded-local-variance,ass:L-smooth-and-bounded} are satisfied and $\eta_\ell \leq \frac{1}{4L(P+1)}$, QAFeL's $k$-th local step satisfies
    \begin{align}
         & \expec{\normsq{y^{s}_{k,p} - z}} \leq 3\expec{\normsq{y_{k,0}^s - z}}                                              \\
         & + 10P^2 \eta_\ell^2 \expec{\sigma_\ell^2 + \normsq{\nabla f(z)} + \normsq{\nabla F_k(z) - \nabla f(z)}}. \nonumber
    \end{align}
\end{lemma}
\begin{IEEEproof}
    We start by expanding the $p$-th local step:
    \begin{equation}
        \normsq{y^{s}_{k,p} - z} = \normsq{y^{s}_{k,p-1} - \eta_\ell g_{k}(y_{k,p-1}^s) - z}.
    \end{equation}
    Next, we add and subtract $\eta_\ell \nabla F_k(y_{k,p-1}^s)$, $\eta_\ell \nabla F_k(z)$, and $\eta_\ell \nabla f(z)$, and apply Cauchy-Schwarz:
    \begin{align*}
        \normsq{y^{s}_{k,p} - z} & \leq (1+c)\normsq{y^{s}_{k,p-1} - z}                                           \\
                                 & + 4\eta_\ell^2(1+c^{-1}) \normsq{g_{k}(y_{k,p-1}^s) - \nabla F_k(y_{k,p-1}^s)} \\
                                 & + 4\eta_\ell^2(1+c^{-1}) \normsq{\nabla F_k(y_{k,p-1}^s) - \nabla F_k(z)}      \\
                                 & + 4\eta_\ell^2(1+c^{-1}) \normsq{\nabla F_k(z) - \nabla f(z)}                  \\
                                 & + 4\eta_\ell^2(1+c^{-1}) \normsq{\nabla f(z)}.
    \end{align*}
    Selecting $c = \frac{1}{2P}$, and applying \cref{ass:bounded-local-variance,ass:L-smooth-and-bounded,ass:bounded-heterogeneity}, we obtain
    \begin{align*}
        \expec{\normsq{y^{s}_{k,p} - z}} & \leq \left ( 1+\frac{1}{2P} \right ) \expec{\normsq{y^{s}_{k,p-1} - z}} \\
                                         & + 4\eta_\ell^2(1+2P) \sigma_\ell^2                                      \\
                                         & + 4\eta_\ell^2(1+2P) L^2 \expec{\normsq{y_{k,p-1}^s - z}}               \\
                                         & + 4\eta_\ell^2(1+2P) \expec{\normsq{\nabla F_k(z) - \nabla f(z)}}       \\
                                         & + 4\eta_\ell^2(1+2P) \normsq{\nabla f(z)}.
    \end{align*}
    Now, using that $\eta_\ell \leq \frac{1}{4L(P+1)} \implies 4\eta_\ell^2(1+2P)L^2 \leq \frac{1}{2P}$, we define the $U_p$ series and $R$ term for ease of notation in the next steps as follows:
    \begin{multline*}
        \underbrace{\expec{\normsq{y^{s}_{k,p} - z}}}_{U_{p}} \leq \left ( 1+\frac{1}{P} \right ) \underbrace{\expec{\normsq{y^{s}_{k,p-1} - z}}}_{U_{p-1}} \\
        {} + \underbrace{4\eta_\ell^2(1+2P) (\sigma_\ell^2 + \expec{\normsq{\nabla F_k(z) - \nabla f(z)} + \normsq{\nabla f(z)}})}_{R}.
    \end{multline*}
    Finally, we use this recursion $p$ times and obtain
    \begin{equation}
        \begin{aligned}
            U_p & \leq U_0 \left ( 1 + \frac{1}{P} \right )^p + R \sum_{i=0}^{p-1} \left ( 1 + \frac{1}{P} \right )^i \\
                & \leq U_0 e + R (e - 1)P,
        \end{aligned}
    \end{equation}
    where we have used the geometric series sum and basic properties of the exponential function to obtain the last inequality.
    To finish the proof, we just apply the definitions of $U_0$ and $R$, and bound $e < 3$, as well as $(e-1)4(1+2P)P < 10P^2$, given that $P$ is no less than 1.
\end{IEEEproof}

\section{Algorithm Pseudocode}\label{app:pseudo}
This appendix contains the pseudocode for all the components of QAFeL, see \cref{alg:server,alg:client,alg:client-background}.
The highlighted lines are the key novelties of the algorithm, which constitute the hidden-state mechanism.
\begin{algorithm}[p]
    \caption{\texttt{QAFeL-server}} \label{alg:server}
    \begin{algorithmic}[1]
        \REQUIRE {server learning rate $\eta_g$, client learning rate $\eta_{\ell}$, client SGD steps $P$, buffer size $K$, initial model $x^0$}
        \STATE \mybox{$\hat x^0 \leftarrow x^0$ \hspace{2cm} \{initialize shared hidden state\} }
        \REPEAT
        \STATE{run {QAFeL-client}$(\eta_{\ell}, P)$ on clients \hfill\COMMENT{async}}
        \IF{receive client update}
        \STATE{$\Delta_n \leftarrow$ received \mybox{quantized} update from client $n$}
        \STATE{$\overline{\Delta}^t \leftarrow \overline{\Delta}^t + \Delta_n$}
        \STATE{$k \leftarrow k + 1$} \hfill\COMMENT{number of clients in buffer}
        \ENDIF
        \IF{$k == K$}
        \STATE {$\overline{\Delta}^t \leftarrow \frac{\overline{\Delta}^t}{K}$}
        \STATE {$x^{t+1} \leftarrow x^t + \eta_g \overline{\Delta}^t$}
        \STATE {\mybox{Broadcast $q^t \leftarrow Q_s(x^{t+1} - \hat x^t)$}}
        \STATE \mybox{$\hat x^{t+1} \leftarrow \hat x^t + q^t$ \hspace{.7cm} \{update shared hidden state\} }
        \STATE  {$\overline{\Delta}^t \leftarrow 0, k \leftarrow 0, t \leftarrow t + 1$ \hfill\COMMENT{reset buffer}}
        \ENDIF
        \UNTIL{convergence}
        \ENSURE {FL-trained global model}
    \end{algorithmic}
\end{algorithm}
\begin{algorithm}[p]
    \caption{\texttt{QAFeL-client}} \label{alg:client}
    \begin{algorithmic}[1]
        \REQUIRE client learning rate $\eta_{\ell}$, number of client SGD steps $P$
        \STATE{$y_0 \leftarrow$ \mybox{$\hat x^t$ \hspace{0.2cm} \{availability of $\hat x^t$ ensured by algorithm~\ref{alg:client-background}\} }}
        \FOR{$p=1:P$}
        \STATE{$y_p \leftarrow y_{p-1} - \eta_{\ell} g_p(y_{p-1} )  $}
        \ENDFOR
        \STATE{$\Delta \leftarrow$ \mybox{$ Q_c(y_0 - y_p)$ \hspace{.8cm} \{Using an ubiased quantizer\}}}
        \STATE{send $\Delta$ to server}
        \ENSURE client update $\Delta$
    \end{algorithmic}
\end{algorithm}
\begin{algorithm}[p]
    \caption{\mybox{\texttt{QAFeL{}-client-background}}} \label{alg:client-background}
    \begin{algorithmic}[1]
        \REQUIRE {initial model $x^0$}
        \STATE {$\hat x^0 \leftarrow x^0$}
        \REPEAT
        \STATE {wait for quantized update $q^t$}
        \STATE {$\hat x^{t+1} \leftarrow \hat x^t + q^t$}
        \UNTIL{shutdown}
        \ENSURE {updated FL global model}
    \end{algorithmic}
\end{algorithm}

\bibliography{references}

\begin{thebibliography}{10}
\providecommand{\url}[1]{#1}
\csname url@samestyle\endcsname
\providecommand{\newblock}{\relax}
\providecommand{\bibinfo}[2]{#2}
\providecommand{\BIBentrySTDinterwordspacing}{\spaceskip=0pt\relax}
\providecommand{\BIBentryALTinterwordstretchfactor}{4}
\providecommand{\BIBentryALTinterwordspacing}{\spaceskip=\fontdimen2\font plus
\BIBentryALTinterwordstretchfactor\fontdimen3\font minus
  \fontdimen4\font\relax}
\providecommand{\BIBforeignlanguage}[2]{{%
\expandafter\ifx\csname l@#1\endcsname\relax
\typeout{** WARNING: IEEEtran.bst: No hyphenation pattern has been}%
\typeout{** loaded for the language `#1'. Using the pattern for}%
\typeout{** the default language instead.}%
\else
\language=\csname l@#1\endcsname
\fi
#2}}
\providecommand{\BIBdecl}{\relax}
\BIBdecl

\bibitem{QAFeLworkshop}
T.~Ortega and H.~Jafarkhani, ``Asynchronous federated learning with
  bidirectional quantized communications and buffered aggregation,'' in
  \emph{2023 ICML Workshop of Federated Learning and Analytics in Practice},
  2023.

\bibitem{D-ADMM}
J.~F.~C. Mota, J.~M.~F. Xavier, P.~M.~Q. Aguiar, and M.~Püschel, ``: A
  communication-efficient distributed algorithm for separable optimization,''
  \emph{IEEE Transactions on Signal Processing}, vol.~61, no.~10, pp.
  2718--2723, 2013.

\bibitem{shen2021distributed}
Y.~Shen, S.~Karimi-Bidhendi, and H.~Jafarkhani, ``Distributed and quantized
  online multi-kernel learning,'' \emph{IEEE Transactions on Signal
  Processing}, vol.~69, pp. 5496--5511, 2021.

\bibitem{chocosgd}
\BIBentryALTinterwordspacing
A.~Koloskova, S.~Stich, and M.~Jaggi, ``\BIBforeignlanguage{en}{Decentralized
  stochastic optimization and gossip algorithms with compressed
  communication},'' in \emph{\BIBforeignlanguage{en}{Proceedings of the 36th
  International Conference on Machine Learning}}.\hskip 1em plus 0.5em minus
  0.4em\relax PMLR, 5 2019, p. 3478–3487. [Online]. Available:
  \url{https://proceedings.mlr.press/v97/koloskova19a.html}
\BIBentrySTDinterwordspacing

\bibitem{communication_efficient}
\BIBentryALTinterwordspacing
B.~McMahan, E.~Moore, D.~Ramage, S.~Hampson, and B.~A.~y. Arcas,
  ``\BIBforeignlanguage{en}{Communication-efficient learning of deep networks
  from decentralized data},'' in \emph{\BIBforeignlanguage{en}{Proceedings of
  the 20th International Conference on Artificial Intelligence and
  Statistics}}.\hskip 1em plus 0.5em minus 0.4em\relax PMLR, 4 2017, p.
  1273–1282. [Online]. Available:
  \url{https://proceedings.mlr.press/v54/mcmahan17a.html}
\BIBentrySTDinterwordspacing

\bibitem{advances_open_problems}
\BIBentryALTinterwordspacing
P.~Kairouz \emph{et~al.}, ``\BIBforeignlanguage{en}{Advances and open problems
  in federated learning},'' \emph{\BIBforeignlanguage{en}{Foundations and
  Trends® in Machine Learning}}, vol.~14, no. 1–2, p. 1–210, 2021.
  [Online]. Available:
  \url{http://www.nowpublishers.com/article/Details/MAL-083}
\BIBentrySTDinterwordspacing

\bibitem{tfl-dt}
\BIBentryALTinterwordspacing
J.~Guo \emph{et~al.}, ``{TFL-DT}: A trust evaluation scheme for federated
  learning in digital twin for mobile networks,'' \emph{IEEE Journal on
  Selected Areas in Communications}, vol.~41, no.~11, pp. 3548--3560, 2023.
  [Online]. Available: \url{https://ieeexplore.ieee.org/document/10234616}
\BIBentrySTDinterwordspacing

\bibitem{joint_TCOM}
\BIBentryALTinterwordspacing
M.~Chen, Z.~Yang, W.~Saad, C.~Yin, H.~V. Poor, and S.~Cui, ``A joint learning
  and communications framework for federated learning over wireless networks,''
  \emph{IEEE Transactions on Wireless Communications}, vol.~20, no.~1, pp.
  269--283, Jan. 2021. [Online]. Available:
  \url{https://ieeexplore.ieee.org/abstract/document/9210812}
\BIBentrySTDinterwordspacing

\bibitem{ML_model_sizes}
\BIBentryALTinterwordspacing
P.~Villalobos, J.~Sevilla, T.~Besiroglu, L.~Heim, A.~Ho, and M.~Hobbhahn,
  ``Machine learning model sizes and the parameter gap,'' \emph{arXiv}, no.
  arXiv:2207.02852, 7 2022, arXiv:2207.02852 [cs]. [Online]. Available:
  \url{http://arxiv.org/abs/2207.02852}
\BIBentrySTDinterwordspacing

\bibitem{fedavg_conv_Li}
\BIBentryALTinterwordspacing
X.~Li, K.~Huang, W.~Yang, S.~Wang, and Z.~Zhang, ``\BIBforeignlanguage{en}{On
  the convergence of fedavg on non-iid data},'' in
  \emph{\BIBforeignlanguage{en}{ICLR 2020}}, 7 2019. [Online]. Available:
  \url{https://arxiv.org/abs/1907.02189v4}
\BIBentrySTDinterwordspacing

\bibitem{fedprox}
\BIBentryALTinterwordspacing
T.~Li, A.~K. Sahu, M.~Zaheer, M.~Sanjabi, A.~Talwalkar, and V.~Smith,
  ``Federated optimization in heterogeneous networks,'' in \emph{MLSys
  2020}.\hskip 1em plus 0.5em minus 0.4em\relax arXiv, 4 2020, arXiv:1812.06127
  [cs, stat]. [Online]. Available: \url{http://arxiv.org/abs/1812.06127}
\BIBentrySTDinterwordspacing

\bibitem{async_hetero}
\BIBentryALTinterwordspacing
C.~Xu, Y.~Qu, Y.~Xiang, and L.~Gao, ``Asynchronous federated learning on
  heterogeneous devices: A survey,'' in \emph{arXiv}, 8 2022, arXiv:2109.04269
  [cs]. [Online]. Available: \url{http://arxiv.org/abs/2109.04269}
\BIBentrySTDinterwordspacing

\bibitem{async_edge}
\BIBentryALTinterwordspacing
Y.~Chen, Y.~Ning, M.~Slawski, and H.~Rangwala, ``Asynchronous online federated
  learning for edge devices with non-iid data,'' in \emph{2020 IEEE
  International Conference on Big Data (Big Data)}.\hskip 1em plus 0.5em minus
  0.4em\relax Los Alamitos, CA, USA: IEEE Computer Society, 12 2020, pp.
  15--24. [Online]. Available:
  \url{https://doi.ieeecomputersociety.org/10.1109/BigData50022.2020.9378161}
\BIBentrySTDinterwordspacing

\bibitem{papaya}
\BIBentryALTinterwordspacing
D.~Huba \emph{et~al.}, ``\BIBforeignlanguage{en}{Papaya: Practical, private,
  and scalable federated learning},'' \emph{\BIBforeignlanguage{en}{Proceedings
  of Machine Learning and Systems}}, vol.~4, p. 814–832, 4 2022. [Online].
  Available:
  \url{https://proceedings.mlsys.org/paper/2022/hash/f340f1b1f65b6df5b5e3f94d95b11daf-Abstract.html}
\BIBentrySTDinterwordspacing

\bibitem{FedBuff}
\BIBentryALTinterwordspacing
J.~Nguyen \emph{et~al.}, ``\BIBforeignlanguage{en}{Federated learning with
  buffered asynchronous aggregation},'' in
  \emph{\BIBforeignlanguage{en}{Proceedings of The 25th International
  Conference on Artificial Intelligence and Statistics}}.\hskip 1em plus 0.5em
  minus 0.4em\relax PMLR, 5 2022, p. 3581–3607. [Online]. Available:
  \url{https://proceedings.mlr.press/v151/nguyen22b.html}
\BIBentrySTDinterwordspacing

\bibitem{toghani2022unbounded}
M.~T. Toghani and C.~A. Uribe, ``Unbounded gradients in federated learning with
  buffered asynchronous aggregation,'' in \emph{2022 58th Annual Allerton
  Conference on Communication, Control, and Computing (Allerton)}.\hskip 1em
  plus 0.5em minus 0.4em\relax IEEE, 2022, pp. 1--8.

\bibitem{robust-async}
Y.~Miao \emph{et~al.}, ``Robust asynchronous federated learning with
  time-weighted and stale model aggregation,'' \emph{IEEE Transactions on
  Dependable and Secure Computing}, pp. 1--15, 2023.

\bibitem{backdoor-FL}
Y.~Miao, R.~Xie, X.~Li, Z.~Liu, K.-K.~R. Choo, and R.~H. Deng, ``Efficient and
  secure federated learning against backdoor attacks,'' \emph{IEEE Transactions
  on Dependable and Secure Computing}, pp. 1--18, 2024.

\bibitem{zhou2022fedaca}
S.~Zhou, Y.~Huo, S.~Bao, B.~Landman, and A.~Gokhale, ``Fedaca: An adaptive
  communication-efficient asynchronous framework for federated learning,'' in
  \emph{2022 IEEE International Conference on Autonomic Computing and
  Self-Organizing Systems (ACSOS)}.\hskip 1em plus 0.5em minus 0.4em\relax
  IEEE, 2022, pp. 71--80.

\bibitem{UVeQFed}
N.~Shlezinger, M.~Chen, Y.~C. Eldar, H.~V. Poor, and S.~Cui, ``{UVeQFed}:
  Universal vector quantization for federated learning,'' \emph{IEEE
  Transactions on Signal Processing}, vol.~69, p. 500–514, 2021.

\bibitem{optimal_compression}
\BIBentryALTinterwordspacing
A.~Albasyoni, M.~Safaryan, L.~Condat, and P.~Richtárik, ``Optimal gradient
  compression for distributed and federated learning,'' in \emph{SpicyFL
  2020}.\hskip 1em plus 0.5em minus 0.4em\relax arXiv, 2020. [Online].
  Available: \url{https://arxiv.org/abs/2010.03246}
\BIBentrySTDinterwordspacing

\bibitem{error_feedback}
\BIBentryALTinterwordspacing
S.~P. Karimireddy, Q.~Rebjock, S.~Stich, and M.~Jaggi,
  ``\BIBforeignlanguage{en}{Error feedback fixes sign{SGD} and other gradient
  compression schemes},'' in \emph{\BIBforeignlanguage{en}{Proceedings of the
  36th International Conference on Machine Learning}}.\hskip 1em plus 0.5em
  minus 0.4em\relax PMLR, 5 2019, p. 3252–3261. [Online]. Available:
  \url{https://proceedings.mlr.press/v97/karimireddy19a.html}
\BIBentrySTDinterwordspacing

\bibitem{unbiased_horvath}
\BIBentryALTinterwordspacing
S.~Horv{\'a}th and P.~Richtarik, ``A better alternative to error feedback for
  communication-efficient distributed learning,'' in \emph{International
  Conference on Learning Representations}, 2021. [Online]. Available:
  \url{https://openreview.net/forum?id=vYVI1CHPaQg}
\BIBentrySTDinterwordspacing

\bibitem{joint_privacy_quantization}
N.~Lang, E.~Sofer, T.~Shaked, and N.~Shlezinger, ``Joint privacy enhancement
  and quantization in federated learning,'' \emph{IEEE Transactions on Signal
  Processing}, vol.~71, p. 295–310, 2023.

\bibitem{qsgd}
\BIBentryALTinterwordspacing
D.~Alistarh, D.~Grubic, J.~Li, R.~Tomioka, and M.~Vojnovic, ``{QSGD}:
  Communication-efficient {SGD} via gradient quantization and encoding,'' in
  \emph{Advances in Neural Information Processing Systems}, I.~Guyon
  \emph{et~al.}, Eds., vol.~30.\hskip 1em plus 0.5em minus 0.4em\relax Curran
  Associates, Inc., 2017. [Online]. Available:
  \url{https://proceedings.neurips.cc/paper/2017/file/6c340f25839e6acdc73414517203f5f0-Paper.pdf}
\BIBentrySTDinterwordspacing

\bibitem{LIBSVM}
\BIBentryALTinterwordspacing
C.-C. Chang and C.-J. Lin, ``\BIBforeignlanguage{en}{{LIBSVM}: A library for
  support vector machines},'' \emph{\BIBforeignlanguage{en}{ACM Transactions on
  Intelligent Systems and Technology}}, vol.~2, no.~3, p. 1–27, 4 2011.
  [Online]. Available: \url{https://dl.acm.org/doi/10.1145/1961189.1961199}
\BIBentrySTDinterwordspacing

\bibitem{error_feedback_video}
B.~Girod and N.~Farber, ``Feedback-based error control for mobile video
  transmission,'' \emph{Proceedings of the IEEE}, vol.~87, no.~10, pp.
  1707--1723, 1999.

\bibitem{video_codec_design}
\BIBentryALTinterwordspacing
I.~E.~G. Richardson, \emph{\BIBforeignlanguage{en}{Video Codec Design:
  Developing Image and Video Compression Systems}}, 1st~ed.\hskip 1em plus
  0.5em minus 0.4em\relax Wiley, 4 2002. [Online]. Available:
  \url{https://onlinelibrary.wiley.com/doi/book/10.1002/0470847832}
\BIBentrySTDinterwordspacing

\bibitem{md-video-coding}
A.~Reibman, H.~Jafarkhani, Y.~Wang, M.~Orchard, and R.~Puri,
  ``Multiple-description video coding using motion-compensated temporal
  prediction,'' \emph{IEEE Transactions on Circuits and Systems for Video
  Technology}, vol.~12, no.~3, pp. 193--204, 2002.

\bibitem{quantization}
R.~Gray and D.~Neuhoff, ``Quantization,'' \emph{IEEE Transactions on
  Information Theory}, vol.~44, no.~6, p. 2325–2383, 10 1998.

\bibitem{sparsifiedSGD}
\BIBentryALTinterwordspacing
S.~U. Stich, J.-B. Cordonnier, and M.~Jaggi, ``Sparsified {SGD} with memory,''
  in \emph{Advances in Neural Information Processing Systems}, vol.~31.\hskip
  1em plus 0.5em minus 0.4em\relax Curran Associates, Inc., 2018. [Online].
  Available:
  \url{https://proceedings.neurips.cc/paper/2018/hash/b440509a0106086a67bc2ea9df0a1dab-Abstract.html}
\BIBentrySTDinterwordspacing

\bibitem{sparsifiedGradientMethods}
\BIBentryALTinterwordspacing
D.~Alistarh, T.~Hoefler, M.~Johansson, N.~Konstantinov, S.~Khirirat, and
  C.~Renggli, ``The convergence of sparsified gradient methods,'' in
  \emph{Advances in Neural Information Processing Systems}, vol.~31.\hskip 1em
  plus 0.5em minus 0.4em\relax Curran Associates, Inc., 2018. [Online].
  Available:
  \url{https://proceedings.neurips.cc/paper_files/paper/2018/hash/314450613369e0ee72d0da7f6fee773c-Abstract.html}
\BIBentrySTDinterwordspacing

\bibitem{EF21}
\BIBentryALTinterwordspacing
P.~Richtarik, I.~Sokolov, and I.~Fatkhullin, ``{EF21}: A new, simpler,
  theoretically better, and practically faster error feedback,'' in
  \emph{Advances in Neural Information Processing Systems}, vol.~34.\hskip 1em
  plus 0.5em minus 0.4em\relax Curran Associates, Inc., 2021, p. 4384–4396.
  [Online]. Available:
  \url{https://proceedings.neurips.cc/paper/2021/hash/231141b34c82aa95e48810a9d1b33a79-Abstract.html}
\BIBentrySTDinterwordspacing

\bibitem{ortegaGossip}
T.~Ortega and H.~Jafarkhani, ``Gossiped and quantized online multi-kernel
  learning,'' \emph{IEEE Signal Processing Letters}, vol.~30, pp. 468--472,
  2023.

\bibitem{MCM}
C.~Philippenko and A.~Dieuleveut, ``Preserved central model for faster
  bidirectional compression in distributed settings,'' \emph{Advances in Neural
  Information Processing Systems}, vol.~34, pp. 2387--2399, 2021.

\bibitem{wang}
\BIBentryALTinterwordspacing
J.~Wang, Q.~Liu, H.~Liang, G.~Joshi, and H.~V. Poor, ``Tackling the objective
  inconsistency problem in heterogeneous federated optimization,'' in
  \emph{Advances in Neural Information Processing Systems}, H.~Larochelle,
  M.~Ranzato, R.~Hadsell, M.~Balcan, and H.~Lin, Eds., vol.~33.\hskip 1em plus
  0.5em minus 0.4em\relax Curran Associates, Inc., 2020, pp. 7611--7623.
  [Online]. Available:
  \url{https://proceedings.neurips.cc/paper_files/paper/2020/file/564127c03caab942e503ee6f810f54fd-Paper.pdf}
\BIBentrySTDinterwordspacing

\bibitem{asyncfl}
\BIBentryALTinterwordspacing
C.~Xie, S.~Koyejo, and I.~Gupta, ``Asynchronous federated optimization,'' in
  \emph{OPT2020: 12th Annual Workshop on Optimization for Machine
  Learning}.\hskip 1em plus 0.5em minus 0.4em\relax arXiv, 12 2020,
  arXiv:1903.03934 [cs]. [Online]. Available:
  \url{http://arxiv.org/abs/1903.03934}
\BIBentrySTDinterwordspacing

\bibitem{complexity}
\BIBentryALTinterwordspacing
Y.~Drori and O.~Shamir, ``\BIBforeignlanguage{en}{The complexity of finding
  stationary points with stochastic gradient descent},'' in
  \emph{\BIBforeignlanguage{en}{Proceedings of the 37th International
  Conference on Machine Learning}}.\hskip 1em plus 0.5em minus 0.4em\relax
  PMLR, 11 2020, p. 2658–2667. [Online]. Available:
  \url{https://proceedings.mlr.press/v119/drori20a.html}
\BIBentrySTDinterwordspacing

\bibitem{LEAF}
\BIBentryALTinterwordspacing
S.~Caldas \emph{et~al.}, ``\BIBforeignlanguage{en}{{LEAF}: A benchmark for
  federated settings},'' in \emph{\BIBforeignlanguage{en}{NeurIPS 2019}}, 12
  2018. [Online]. Available: \url{https://arxiv.org/abs/1812.01097v3}
\BIBentrySTDinterwordspacing

\bibitem{CIFAR}
\BIBentryALTinterwordspacing
A.~Krizhevsky, G.~Hinton \emph{et~al.}, ``Learning multiple layers of features
  from tiny images,'' \emph{University of Toronto, ON, Canada}, 2009. [Online].
  Available:
  \url{https://www.cs.toronto.edu/~kriz/learning-features-2009-TR.pdf}
\BIBentrySTDinterwordspacing

\bibitem{loss-multilayer}
\BIBentryALTinterwordspacing
A.~Choromanska, M.~Henaff, M.~Mathieu, G.~Ben~Arous, and Y.~LeCun, ``The loss
  surfaces of multilayer networks,'' in \emph{Proceedings of the Eighteenth
  International Conference on Artificial Intelligence and Statistics}, ser.
  Proceedings of Machine Learning Research, G.~Lebanon and S.~V.~N.
  Vishwanathan, Eds., vol.~38.\hskip 1em plus 0.5em minus 0.4em\relax San
  Diego, California, USA: PMLR, 5 2015, pp. 192--204. [Online]. Available:
  \url{https://proceedings.mlr.press/v38/choromanska15.html}
\BIBentrySTDinterwordspacing

\bibitem{Nesterov_2018}
\BIBentryALTinterwordspacing
Y.~Nesterov, \emph{Lectures on Convex Optimization}, ser. Springer Optimization
  and Its Applications.\hskip 1em plus 0.5em minus 0.4em\relax Cham: Springer
  International Publishing, 2018, vol. 137. [Online]. Available:
  \url{http://link.springer.com/10.1007/978-3-319-91578-4}
\BIBentrySTDinterwordspacing

\bibitem{balancing_TCOM}
M.~K. Nori, S.~Yun, and I.-M. Kim, ``Fast federated learning by balancing
  communication trade-offs,'' \emph{IEEE Transactions on Communications},
  vol.~69, no.~8, pp. 5168--5182, 2021.

\bibitem{tighter_theory}
\BIBentryALTinterwordspacing
A.~Khaled, K.~Mishchenko, and P.~Richtarik, ``\BIBforeignlanguage{en}{Tighter
  theory for local {SGD} on identical and heterogeneous data},'' in
  \emph{\BIBforeignlanguage{en}{Proceedings of the Twenty Third International
  Conference on Artificial Intelligence and Statistics}}.\hskip 1em plus 0.5em
  minus 0.4em\relax PMLR, 6 2020, p. 4519–4529. [Online]. Available:
  \url{https://proceedings.mlr.press/v108/bayoumi20a.html}
\BIBentrySTDinterwordspacing

\bibitem{flsim}
\BIBentryALTinterwordspacing
L.~Li, J.~Wang, and C.~Xu, ``Flsim: An extensible and reusable simulation
  framework for federated learning,'' in \emph{Simulation Tools and
  Techniques}, H.~Song and D.~Jiang, Eds.\hskip 1em plus 0.5em minus
  0.4em\relax Cham: Springer International Publishing, 2021, pp. 350--369.
  [Online]. Available: \url{https://doi.org/10.1007/978-3-030-72792-5_30}
\BIBentrySTDinterwordspacing

\bibitem{pytorch}
\BIBentryALTinterwordspacing
A.~Paszke \emph{et~al.}, ``Pytorch: An imperative style, high-performance deep
  learning library,'' in \emph{Advances in Neural Information Processing
  Systems}, vol.~32.\hskip 1em plus 0.5em minus 0.4em\relax Curran Associates,
  Inc., 2019. [Online]. Available:
  \url{https://proceedings.neurips.cc/paper/2019/hash/bdbca288fee7f92f2bfa9f7012727740-Abstract.html}
\BIBentrySTDinterwordspacing

\bibitem{celeba}
\BIBentryALTinterwordspacing
Z.~Liu, P.~Luo, X.~Wang, and X.~Tang, ``Deep learning face attributes in the
  wild,'' in \emph{2015 IEEE International Conference on Computer Vision
  (ICCV)}.\hskip 1em plus 0.5em minus 0.4em\relax Los Alamitos, CA, USA: IEEE
  Computer Society, 12 2015, pp. 3730--3738. [Online]. Available:
  \url{https://doi.ieeecomputersociety.org/10.1109/ICCV.2015.425}
\BIBentrySTDinterwordspacing

\bibitem{dirichlet}
\BIBentryALTinterwordspacing
T.-M.~H. Hsu, H.~Qi, and M.~Brown, ``Measuring the effects of non-identical
  data distribution for federated visual classification,'' \emph{arXiv}, no.
  arXiv:1909.06335, 9 2019. [Online]. Available:
  \url{http://arxiv.org/abs/1909.06335}
\BIBentrySTDinterwordspacing

\bibitem{groupnorm}
\BIBentryALTinterwordspacing
Y.~Wu and K.~He, ``\BIBforeignlanguage{en}{Group normalization},'' in
  \emph{\BIBforeignlanguage{en}{Computer Vision – ECCV 2018}}, ser. Lecture
  Notes in Computer Science, V.~Ferrari, M.~Hebert, C.~Sminchisescu, and
  Y.~Weiss, Eds.\hskip 1em plus 0.5em minus 0.4em\relax Cham: Springer
  International Publishing, 2018, p. 3–19. [Online]. Available:
  \url{https://doi.org/10.1007/978-3-030-01261-8}
\BIBentrySTDinterwordspacing

\bibitem{batchnorm_with_groupnorm}
\BIBentryALTinterwordspacing
K.~Hsieh, A.~Phanishayee, O.~Mutlu, and P.~Gibbons,
  ``\BIBforeignlanguage{en}{The non-iid data quagmire of decentralized machine
  learning},'' in \emph{\BIBforeignlanguage{en}{Proceedings of the 37th
  International Conference on Machine Learning}}.\hskip 1em plus 0.5em minus
  0.4em\relax PMLR, 11 2020, p. 4387–4398. [Online]. Available:
  \url{https://proceedings.mlr.press/v119/hsieh20a.html}
\BIBentrySTDinterwordspacing

\bibitem{bayesian}
\BIBentryALTinterwordspacing
J.~Snoek, H.~Larochelle, and R.~P. Adams, ``Practical bayesian optimization of
  machine learning algorithms,'' in \emph{Advances in Neural Information
  Processing Systems}, F.~Pereira, C.~Burges, L.~Bottou, and K.~Weinberger,
  Eds., vol.~25.\hskip 1em plus 0.5em minus 0.4em\relax Curran Associates,
  Inc., 2012. [Online]. Available:
  \url{https://proceedings.neurips.cc/paper_files/paper/2012/file/05311655a15b75fab86956663e1819cd-Paper.pdf}
\BIBentrySTDinterwordspacing

\end{thebibliography}
\bibliographystyle{IEEEtran}

\begin{IEEEbiography}[{\includegraphics[width=1in,height=1.25in,clip,keepaspectratio]{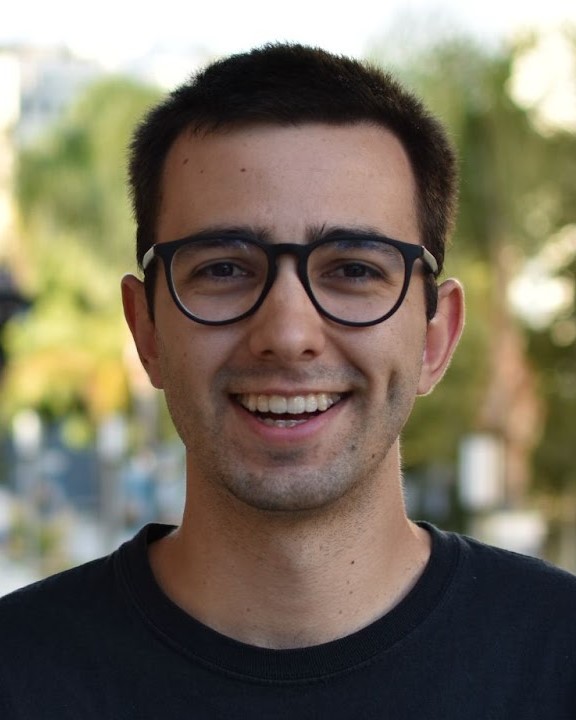}}]{Tomas Ortega} (Graduate Student Member, IEEE) is a PhD candidate at the University of California, Irvine. He was previously a JPL Graduate Fellow at the Communication Architectures and Research Section (332). He holds bachelor's degrees in both telecommunications engineering and mathematics, as well as master's degree in Advanced Mathematics and Mathematical Engineering from Universitat Politecnica de Catalunya, Barcelona. He is a recipient of a 2023 IEEE Signal Processing Scholarship, as well as a Balsells Fellowship and the University of California Irvine Graduate Fellowship.
\end{IEEEbiography}

\begin{IEEEbiography}[{\includegraphics[width=1in,height=1.25in,clip,keepaspectratio]{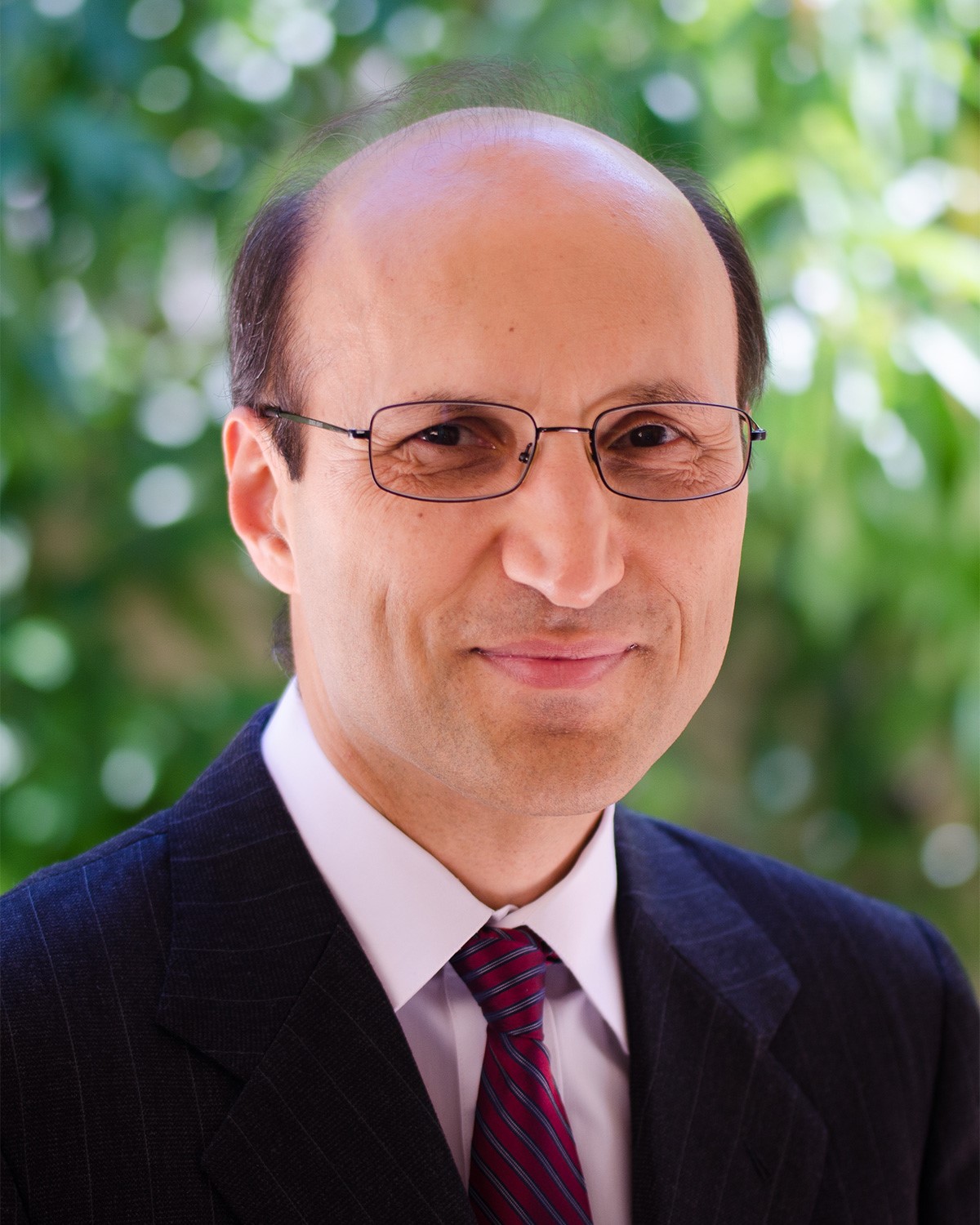}}]{Hamid Jafarkhani} (Fellow, IEEE) is a Chancellor's Professor at the Department of Electrical Engineering and Computer Science, University of California, Irvine, where he is also the Director of Center for Pervasive Communications and Computing,  the former Director of Networked Systems Program, and the Conexant-Broadcom Endowed Chair. He was a Visiting Scholar at Harvard University in 2015, a Visiting Professor at California Institute of Technology in 2018, and a Visiting Research Scholar at Duke University in 2023. He was the 2020-2022 elected Faculty Chair of the UCI School of Engineering. 

Among his awards are the NSF Career Award, the UCI Distinguished Mid-Career Faculty Award for Research, the School of Engineering Excellence in Research Senior Career Award,  the IEEE Marconi Prize Paper Award in Wireless Communications, the IEEE Communications Society Award for Advances in Communication, the IEEE Wireless Communications Technical Committee Recognition Award, the IEEE Signal Processing and Computing for Communications Technical Recognition Award, couple of conference best paper awards, and the IEEE Eric E. Sumner Award. He is the 2017 Innovation Hall of Fame Inductee at the University of Maryland's School of Engineering. 

He was an Associate Editor for the IEEE Communications Letters from 2001-2005, an editor for the IEEE Transactions on Wireless Communications from 2002-2007, an editor for the IEEE Transactions on Communications from 2005-2007, an area
editor for the IEEE Transactions on Wireless Communications from 2007-2012, and 
a Steering Committee Member of the IEEE Transactions on Wireless communications from 2013-2016. He was the general chair of the 2015 IEEE Communication Theory Workshop and the general co-chair of the 2018 IEEE Global Conference on Signal and Information Processing (GlobalSIP). He was an IEEE ComSoc Distinguished lecturer.

Dr. Jafarkhani is listed as an ISI highly cited researcher. According to the Thomson Scientific, he is one of the top 10 most-cited researchers in the field of ``computer science'' during 1997-2007. He is a Fellow of AAAS, an IEEE Fellow, and the author of the book ``Space-Time Coding: Theory and Practice.''
\end{IEEEbiography}
\vfill

\end{document}